\documentclass[a4paper,11pt]{article}
\usepackage{iftex}
\ifPDFTeX
  \usepackage[utf8]{inputenc}
  \usepackage[T1]{fontenc}
  \usepackage{newtxtext}
  \usepackage{CJKutf8}
  \newcommand{\cjktext}[1]{\begin{CJK}{UTF8}{gbsn}#1\end{CJK}}
\else
  \usepackage{fontspec}
  \IfFontExistsTF{Times New Roman}
    {\setmainfont{Times New Roman}}
    {\setmainfont{TeX Gyre Termes}}
  \newfontfamily\cjkfont{Noto Serif CJK SC}
  \newcommand{\cjktext}[1]{{\cjkfont #1}}
\fi
\usepackage[english]{babel}
\usepackage{geometry}
\usepackage{graphicx}
\usepackage{float}
\usepackage{booktabs}
\usepackage[table,xcdraw]{xcolor} %
\usepackage{fancybox}
\usepackage{amssymb,amsfonts,amsmath,amsthm,amscd}
\usepackage{upgreek}
\usepackage{textcomp}
\usepackage{tikz}
\usepackage{fancyhdr}
\usepackage{setspace}
\usepackage[shortlabels]{enumitem}
\usepackage{indentfirst}
\usepackage{tabularx}
\usepackage{longtable}
\usepackage{tabularray}
\usepackage{adjustbox}
\usepackage{caption}
\usepackage{subcaption}
\usepackage{microtype}
\usepackage{tocloft}
\usepackage{csquotes}
\usepackage{icomma}
\usepackage{listings}
\usepackage[nohyperlinks]{acronym}
\usepackage[nottoc,numbib]{tocbibind}
\usepackage[numbers]{natbib}
\usepackage[bookmarksnumbered,hyperfootnotes=false,hidelinks]{hyperref}
\usepackage{url}
\fancypagestyle{plain}{%
  \fancyhf{}
  \fancyhead[R]{\thepage}

}

\definecolor{codegreen}{rgb}{0,0.6,0}
\definecolor{codegray}{rgb}{0.5,0.5,0.5}
\definecolor{codepurple}{rgb}{0.58,0,0.82}
\definecolor{backcolour}{rgb}{0.95,0.95,0.92}
\lstdefinestyle{Python}{
  language=Python,
  backgroundcolor=\color{backcolour},
  commentstyle=\color{codegreen},
  keywordstyle=\color{magenta},
  numberstyle=\tiny\color{codegray},
  stringstyle=\color{codepurple},
  basicstyle=\ttfamily\footnotesize,
  breakatwhitespace=false,
  breaklines=true,
  captionpos=b,
  keepspaces=true,
  numbers=left,
  numbersep=5pt,
  showspaces=false,
  showstringspaces=false,
  showtabs=false,
  tabsize=2
}

\title{VERS: Steering Reasoning Models by Editing Their Activation Explanations}
\author{Ihor Vitenko\\\texttt{igor.vitenko13@gmail.com}}
\date{}
\hypersetup{pdftitle={VERS: Steering Reasoning Models by Editing Their Activation Explanations},pdfauthor={Ihor Vitenko}}
\begin{document}
\maketitle
\begin{abstract}
Reasoning language models write a chain of thought before they answer, and
many checks in it do not change the final answer. Natural language
autoencoders (NLAs) explain a model's activations in text and rebuild
activations from text, and editing an explanation can steer a model at a
single token. We study whether such edits can also steer how a model reasons.
We build VERS (Verbalize, Edit, Reconstruct, Steer): it turns explanation edits
at the doubt points of one reasoning trace into one fixed direction, and adds
this direction at paragraph breaks. On the full GSM8K test, VERS cuts generated
tokens by 45.8\% at an accuracy change of $-0.82$ percentage points (pp); on
MATH-500 it saves 31.3\% at $-3.05$~pp. At half strength, its trade-off between
tokens and accuracy lies between those of the published methods SEAL and DEER.
With fitted affine maps and a strength tuned for each model, the direction
transfers to larger Qwen3 models, up to 235B-A22B, and saves 30--35\% of
tokens. On competition problems, however, VERS loses 3--7~pp at every model
size, and a fixed direction did not separate necessary checks from unneeded
ones.
\end{abstract}
\section{Introduction}
\label{sec:introduction}
Reasoning models often check their own work many times. Some checks repair
mistakes; others add length without changing the answer. Related work finds
such \emph{decorative} reasoning across model sizes \cite{ahamoments2025}.
We want a control that removes useless checks but keeps necessary ones.

\begin{figure}[t]
\centering
\includegraphics[width=\linewidth]{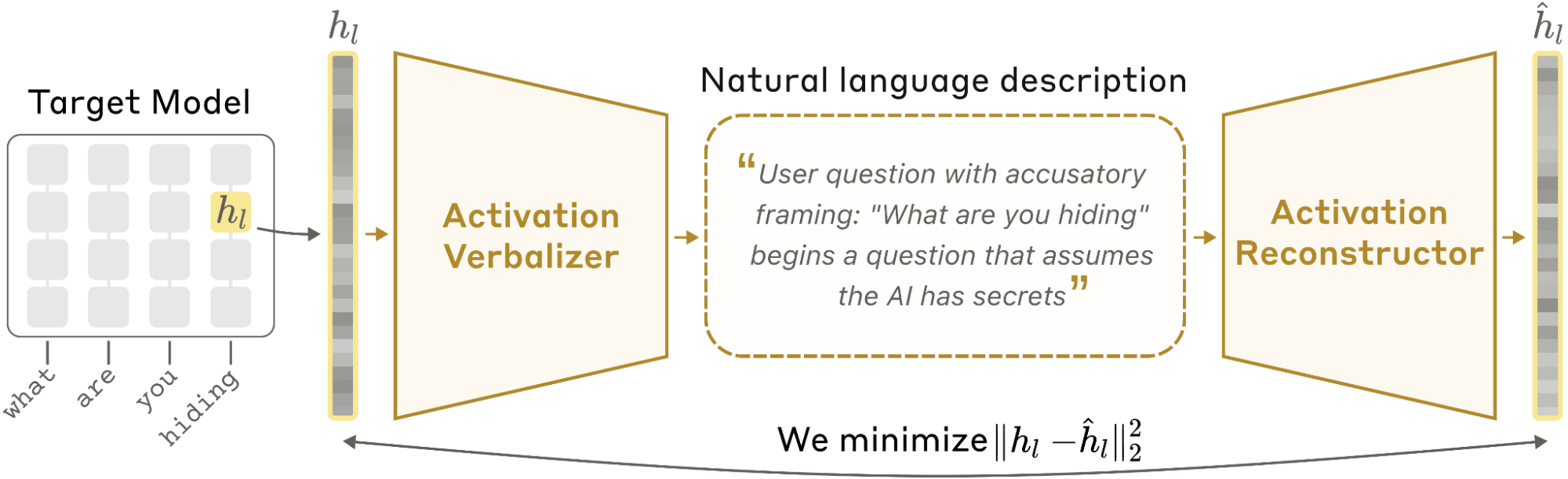}
\caption{A natural language autoencoder. The AV turns the layer-$l$
activation $h_l$ of the target model into a text description. The AR reads the
description and rebuilds the activation as $\hat h_l$.~\cite{frasertaliente2026nla}}
\label{fig:nla}
\end{figure}

Steering directions are usually built from a contrastively labeled corpus:
many examples with the behavior and many without it
\cite{panickssery2023caa,venhoff2025steering}. We ask whether a text edit
can specify the change instead. Natural language autoencoders (NLAs) make this
possible \cite{frasertaliente2026nla}. An NLA is a pair of models trained on
one layer of a language model. The activation verbalizer (AV) reads an
activation and writes a text explanation of it. The activation reconstructor
(AR) reads that explanation and rebuilds the activation
(Figure~\ref{fig:nla}).

The original NLA work already uses explanation edits for interventions
\cite{frasertaliente2026nla}. It edits the explanation of one token, for
example a planned rhyme word ``rabbit'' to ``mouse''. The AR rebuilds both
texts, and the difference of the two rebuilt activations is added at that same
token. One test averages this difference over five explanation samples of the
token. The goal there is to show that explanations are causally linked to
outputs. We study whether such edits can also steer how a model reasons on
math problems. To test this, we build VERS
(\emph{Verbalize, Edit, Reconstruct, Steer}). Compared with the original use,
it adds four things: (a) one
fixed direction, built from one trace, is reused on other questions; (b) it is
added again and again during reasoning, at paragraph breaks found while the
model generates; (c) it is transferred to larger models; and (d) we test where
this control loses selectivity.

In VERS, the AV explains the activation at points where the model is about to
start a check. We replace the predicted check with a continuation of the
solution (the \emph{continue edit}). The AR rebuilds the activation from both
texts, and we average the differences into one fixed direction, $N$. Our
results show that this gives an effective direction built through an NLA
interface. They do not show that the activation-conditioned explanation is
needed: we did not test generic
hand-written texts without AV explanations, or the edited predictions without
the context that the AV describes. Shorter reasoning is our test case. We
measure output length and \emph{detected doubt paragraphs}, whose first
sentence has an English doubt marker such as ``Wait''
(Section~\ref{construction:evidence}). Our three contributions are:
\begin{enumerate}[leftmargin=*,nosep]
\item \textbf{A steering policy built from explanation edits.} Edits at 35
sites in one trace give a fixed direction. On the full GSM8K test (1,319 questions), it saves
45.8\% of tokens at a 0.82~pp accuracy cost. Section~\ref{inference:frozen} also compares $N$ with a
difference-of-means direction built without the NLA.
\item \textbf{Transfer of the interface and the intervention.} Fitted affine
maps let the frozen 1.7B NLA read and rebuild 8B and 30B-A3B states, and carry
$N$ to larger Qwen3 models, up to 235B-A22B. With strength tuned for each model, $N$ saves 30--35\% of tokens on 400 MATH test
questions at all four model sizes.
\item \textbf{Limits of control.} On competition problems, accuracy falls by
3--7~pp at all model sizes. A probe predicts at each paragraph boundary
whether a useful check (one that turns a wrong answer into a right one) starts
next, with AUC 0.84. But the two tested selective constructions did not
improve the accuracy--length trade-off. The opposite
direction $-N$ adds detected doubt paragraphs, but a direction built from
the doubt edit does not.
\end{enumerate}
Section~\ref{sec:method} defines VERS, and Sections~\ref{sec:inference}--\ref{sec:limits}
test the three contributions. Section~\ref{sec:distillation} adds a supporting
study of students fine-tuned on steered traces.

\section{Related Work}
\label{sec:related}
\paragraph{Language interfaces to activations.}
NLA gives the training recipe for the verbalizer and the reconstructor
\cite{frasertaliente2026nla}. It also steers by adding, at the one explained
token, the difference of the AR outputs for the edited and original
explanation. Patchscopes is another language
interface to hidden representations \cite{ghandeharioun2024patchscopes}. We
reuse NLA's edit-and-rebuild step and extend it into a reusable policy
(Section~\ref{sec:introduction}).

\paragraph{Activation steering.}
Activation steering adds a direction to the residual stream at inference time.
ActAdd builds the direction from the activation difference of a pair of
prompts \cite{turner2023actadd}. Contrastive activation addition (CAA)
averages this difference over many contrastive examples
\cite{panickssery2023caa}. Representation engineering reads and controls
concepts through directions in the representation space \cite{zou2023repe}.
VERS also adds a fixed direction, but it builds the direction from text edits
instead of contrastive examples.

\paragraph{Steering and shortening reasoning.}
Reasoning models often think longer than needed, and this \emph{overthinking}
has been measured directly \cite{chen2024overthinking}. Budget forcing in s1
controls thinking length by ending thinking early or by appending ``Wait''
\cite{muennighoff2025s1}. Other work shortens chain of thought by steering
activations. ASC adds the mean difference between the states of long and short
chains of thought \cite{azizi2026asc}. Manifold Steering projects a reflection
direction onto a low-dimensional subspace of the model's states to remove noise
\cite{manifoldsteering2025}. Venhoff et al.\ steer uncertainty and backtracking
with difference-of-means vectors \cite{venhoff2025steering}. Our comparison
direction $D$ is built in the same way as these directions, from the mean
difference between two kinds of states, and it does not use the NLA. SEAL uses
reflection and execution directions \cite{chen2025seal}. Among the published
methods, it is the closest to Manifold Steering, so we use it as the comparison
from this family. NoWait suppresses thinking-related tokens
\cite{wang2025nowait}. DEER exits thinking early when a trial answer is
confident \cite{yang2025deer}. We run these methods with the same questions,
model, and token budget as VERS (Appendix~\ref{appendix:comparators}), and we count DEER's discarded trial-answer
tokens in its cost. Caldarella et al.\ use the gold answer to find the first
correct answer in a trace \cite{caldarella2026thinkingpast}. This motivates
our study of what happens after an answer appears, but it does not give a
stopping rule that can run during generation. Our final policy does not need
to detect answers.

\paragraph{Reuse across models and through training.}
Model stitching, transfer of activation interventions, and steering large
models with small models are precedents for affine transport
\cite{chen2025stitching,oozeer2025interventiontransfer,bello2025lrt}.
A related NLA study maps another model's sparse-autoencoder features into a
verbalizer \cite{fzaffino2026foreignnla}. We test reuse for two separate
goals: reading and rebuilding target-model states, and carrying one steering
vector to the target. The supporting student study uses supervised
fine-tuning and direct preference optimization \cite{rafailov2023dpo}. It also
trains on matched natural traces, to measure what the steered data adds.

\section{VERS: Verbalize, Edit, Reconstruct, Steer}
\label{sec:method}
\begin{figure}[htbp]
\centering
\begin{tikzpicture}[every node/.style={align=center,font=\small}]
\node[draw,rounded corners,text width=2.2cm,minimum height=1.1cm] (av) at (0,0) {Verbalize\\$e_i=\mathrm{AV}(h_i)$};
\node[draw,rounded corners,text width=2.2cm,minimum height=1.1cm] (edit) at (3,0) {Edit\\$e_i\mapsto T(e_i)$};
\node[draw,rounded corners,text width=3.0cm,minimum height=1.1cm] (ar) at (6.4,0) {Reconstruct\\$\delta_i=\mathrm{AR}(T(e_i))$\\${}-{\mathrm{AR}(e_i)}$};
\node[draw,rounded corners,text width=2.8cm,minimum height=1.1cm] (n) at (10.1,0) {Average\\$N\propto\sum_i \delta_i/\|\delta_i\|_2$\\$\|N\|_2=1$};
\draw[->,thick] (av) -- (edit);
\draw[->,thick] (edit) -- (ar);
\draw[->,thick] (ar) -- (n);
\node[draw,rounded corners,text width=8cm,minimum height=0.9cm] (steer) at (5,-2) {Steer new generations at eligible boundaries\\$h_t\leftarrow h_t+\alpha\lVert h_t\rVert_2 N$};
\draw[->,thick] (n.south) -- (10.1,-2) -- (steer.east);
\end{tikzpicture}
\caption{The two parts of VERS. Top: offline, we build the direction $N$ from
the donor sites (Section~\ref{construction:evidence}). Bottom: during
generation, we add $N$ at eligible paragraph boundaries. Generation calls
neither the AV nor the AR.}
\label{fig:method}
\end{figure}
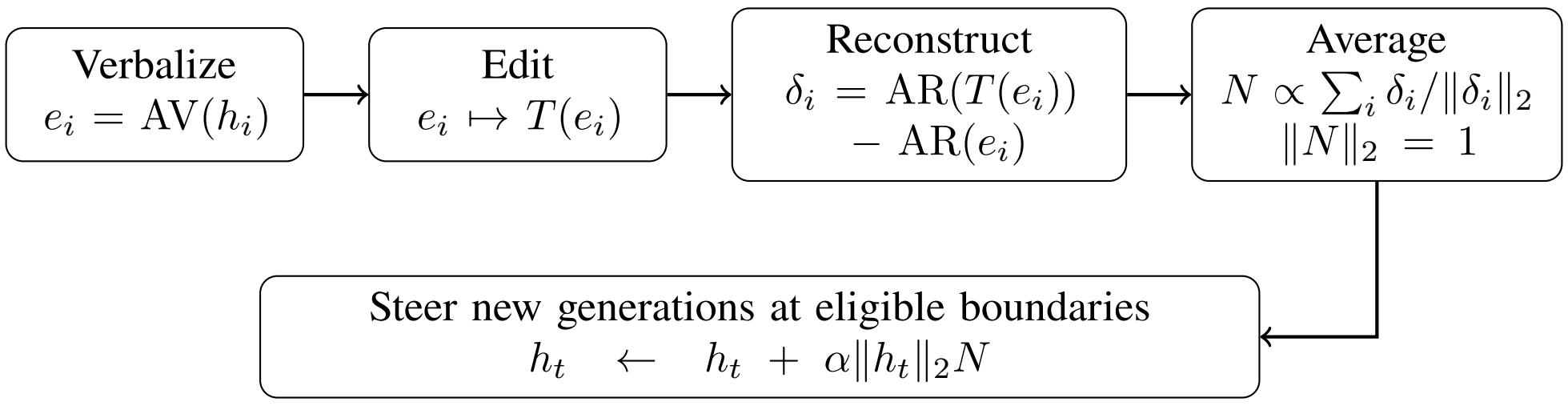

VERS has two parts (Figure~\ref{fig:method}). Offline, it builds one fixed
steering direction, $N$, from hidden states of one trace. The AV turns each
state into text, we edit the text, and the AR maps both texts back to vectors.
During generation, VERS only adds $N$ to the hidden state at paragraph
boundaries. $N$ stands for the NLA direction. Section~\ref{sec:inference}
compares it with $D$, a difference-of-means direction built without the NLA.

\subsection{The trained activation interface}
\label{method:interface}
We train an NLA on Qwen3-1.7B with the training procedure of the original NLA
work \cite{frasertaliente2026nla}, at a smaller scale, and then keep it fixed. It reads the residual
stream at the output of decoder layer 20 (about 70\% of the model's depth, as in the original NLA work \cite{frasertaliente2026nla}). This
state has 2,048 dimensions. The NLA has two parts:
\begin{itemize}
\item The \emph{activation verbalizer} (AV) turns a state into text. It is a full
copy of the model. Its prompt contains one \texttt{<|image\_pad|>} token. We
replace the embedding of this token with the state, scaled to norm 1,000.
Source states have a norm of about 900 on average. Following the original NLA
work \cite{frasertaliente2026nla}, we pick a round number slightly above this
average. The AV then writes an
\emph{explanation} $e$.
\item The \emph{activation reconstructor} (AR) turns text back into a state. It is
the model cut after layer 20, without the final normalization and the
language-model head. A bias-free linear map predicts the state from the
explanation's final token.
\end{itemize}
Neither part changes during the steering and transfer experiments.

Training has two stages. A supervised warm start uses about 500k positions
from Ultra-FineWeb documents, with text summaries written by Qwen3.8-27B. Joint training then uses the reference
reconstruction reward and the GRPO recipe \cite{shao2024deepseekmath} on about
400k new web and chat positions. Appendix~\ref{sec:evidence}
gives the data details.

We measure reconstruction with the fraction of variance explained (FVE). For
$n$ evaluated states $h_j$ with reconstructions $\hat h_j$, we scale both to unit
length, $u_j=h_j/\|h_j\|_2$ and $\hat u_j=\hat h_j/\|\hat h_j\|_2$, and compute
\begin{equation}
\mathrm{FVE}=1-\frac{\sum_{j=1}^{n}\|\hat u_j-u_j\|_2^2}{\sum_{j=1}^{n}\|u_j-\bar u\|_2^2},
\qquad \bar u=\frac{1}{n}\sum_{j=1}^{n}u_j .
\label{eq:fve}
\end{equation}
The mean $\bar u$ is not scaled back to unit length, as in the reference
code \cite{frasertaliente2026nla}. FVE is 1 for a perfect reconstruction. It is 0
when the reconstruction error equals the spread of the scaled states around
their mean $\bar u$. We sample three explanations per state at
temperature 1 and average the FVE over the three samples. The AR reaches an FVE of 0.50
after the warm start and 0.63 after one epoch of joint training. We use this
checkpoint everywhere. On 10,000 new web and chat positions, the FVE is 0.632
(web 0.655, chat 0.604). On 456 positions from reasoning traces, it is only
0.304. So the interface rebuilds reasoning states much less well than web and
chat states. For comparison, the original NLA work reports 0.6--0.8 for its
models \cite{frasertaliente2026nla}.

\subsection{The donor trace and donor sites}
\label{construction:evidence}
We build $N$ from states of one trace. We use these terms:
\begin{itemize}
\item A \emph{paragraph boundary} is a generated token inside thinking (after
\texttt{<think>} and before the first \texttt{</think>}) whose text contains at
least two newline characters, such as \texttt{\textbackslash n\textbackslash n}
or \texttt{.\textbackslash n\textbackslash n}. This token ends a paragraph.
\item A \emph{doubt marker} is one of 13 words or phrases, such as ``Wait'',
``Hmm'', ``But'', and ``Let me verify'' (full list in
Appendix~\ref{sec:evidence}). Matching ignores case.
\item A \emph{doubt paragraph} is a paragraph inside thinking, other than the
first one, whose first sentence contains a doubt marker as a whole word.
Unless we say otherwise, all doubt counts use this rule.
\item The \emph{donor trace} is a correct 1.7B rollout of AIME 2025 I,
Problem 2. We drew 42 correct traces at random, 14 each from GSM8K, AMC 2023,
and AIME 2025, keeping only traces with at least six donor sites. From these, we
used the trace with the most donor sites.
\item A \emph{donor site} is a paragraph boundary in the donor trace where the
next paragraph \emph{starts with} a doubt marker.
\end{itemize}
So a donor site needs a marker at the very start of the next paragraph, and a
doubt paragraph needs one anywhere in its first sentence.

This count is a proxy. It counts English markers only, so it misses checks
without a marker and checks in other languages. In one example, VERS still
checks without a marker, and a ban on doubt words makes the model check in
Chinese (Appendix~\ref{appendix:ellipse}). At full strength on hard problems,
the 1.7B switches into Chinese in 37\% of rollouts (Section~\ref{limits:hard}).

The donor trace has 35 donor sites. At each one we record the layer-20 state
$h_{20}(t)\in\mathbb{R}^{2048}$. Figure~\ref{construction:trace} shows two of
them.

Construction uses information that generation does not have. We chose a trace
with a correct answer, and we chose sites by reading the paragraph that follows.
The steering policy (Section~\ref{method:policy}) uses neither: it steers at
every eligible paragraph boundary, whether or not doubt follows.

\begin{figure}[H]
\centering
\fbox{\begin{minipage}{0.94\linewidth}
\small
\textbf{Donor trace: \texttt{aime2025:1}, rollout 1.}
The problem gives quadrilateral $DEGF$ area 288 and asks for a heptagon's area.
The correct final answer is 588.
\medskip

[\ldots] Take absolute value and multiply by 1/2: 1/2 * |1248| = 624\textcolor{red}{\underline{\texttt{\textbackslash n\textbackslash n}}}
\hfill $t=1809\;\longrightarrow\;h_{20}(1809)$
\medskip

\textbf{But} the problem says that quadrilateral DEGF has area 288. But according
to my calculation, it's 624. Which is a contradiction. Therefore, my coordinates
must be wrong\textcolor{red}{\underline{\texttt{.\textbackslash n\textbackslash n}}}
\hfill $t=1851\;\longrightarrow\;h_{20}(1851)$
\medskip

\textbf{Wait}, that's a problem. So either my coordinates are incorrect, or my
application of shoelace formula is wrong. [\ldots]
\end{minipage}}
\caption{Two of the 35 donor sites. The captured token is underlined in red;
$t$ is its index in the response, counted from zero. The bold words ``But'' and
``Wait'' make these boundaries donor sites; they are not captured. [\ldots]
marks omitted text. Here the doubt is useful: it finds a wrong intermediate
result (624 instead of the given 288).}
\label{construction:trace}
\end{figure}

\subsection{The edit and the direction}
\label{method:edit}
At each donor site $i$, the AV reads the state $h_i$ and writes an explanation
$e_i$. An explanation has several paragraphs, separated by blank lines. The
last paragraph predicts what text comes next. The edit $T$ keeps every paragraph
except the last one and replaces the last one with a fixed template:
\begin{quote}
\textbf{Continue:} This ends the sentence, implying the next token likely
starts a new sentence or clause, such as ``Therefore, the total is calculated
as follows.'' to continue the derivation.

\textbf{Doubt:} This indicates the sentence is concluding the thought
about verification, likely expecting a concluding clause like ``Wait, let me
double-check that.'' to complete the reasoning.
\end{quote}
We call the first edit the \emph{continue edit} and the second the \emph{doubt
edit}. VERS uses the continue edit. Section~\ref{limits:reverse} uses the doubt
edit to build a second direction, $P$. Figure~\ref{fig:explanation} shows a
real explanation and its edited version.

Both templates follow how the AV itself writes (Table~\ref{method:avwords}).
In 289 correct traces, we verbalized three boundaries at a similar depth: one
before a paragraph that starts with ``Wait'', one before another doubt marker,
and one before a paragraph with no marker. Before doubt, the AV names the coming
marker and writes about the answer and checking it. Before a plain paragraph,
it writes about going on to the next step. The doubt template is copied from one
such explanation before ``Wait''. So the continue edit replaces a prediction of
doubt with the kind of prediction the AV already writes before an ordinary step.

The ``answer'' row (shaded in Table~\ref{method:avwords}) shows a pattern. Before
a doubt paragraph, most explanations (58--62\%) say that the model already has
an answer. Before a plain paragraph, only 22.5\% do. So the model tends to start
doubting once it has an answer, and this does not depend on whether the answer
is right. In 1,000 other 1.7B traces (Section~\ref{limits:selectivity}), the
forced answer is already right before 42.6\% of doubt paragraphs, and the check
leaves it unchanged in 97\% of these cases.\footnote{We did not test this, but
we suspect the pattern comes from RL-based post-training
\cite{qwen2025qwen3}.}

\begin{table}[!htb]\centering\small
\begin{tabular}{lrrr}\toprule
& \multicolumn{3}{c}{Next paragraph starts with}\\\cmidrule(lr){2-4}
Word in explanation & ``Wait'' & other doubt marker & no marker\\\midrule
``Wait'' & 43.6\% & 12.1\% & 1.7\%\\
\rowcolor{black!10}``answer'' & 61.6\% & 58.1\% & 22.5\%\\
``check'' & 46.7\% & 53.3\% & 20.8\%\\
``correct'' & 41.2\% & 37.4\% & 8.7\%\\
``continuation'' & 10.0\% & 12.8\% & 32.9\%\\
``next'' & 18.7\% & 20.1\% & 41.5\%\\\bottomrule
\end{tabular}
\caption{Share of AV explanations that contain each word (whole word, any case), at three boundaries
of similar depth in the same 289 correct traces. Each column is a separate set
of 289 explanations. One explanation can contain several words, so the values do
not sum to 100\%. Paired McNemar tests against
the plain boundary give $p<10^{-6}$ for every row (for ``Wait'', the boundary
before ``Wait''; for the other words, the boundary before another doubt
marker).}
\label{method:avwords}
\end{table}

\begin{figure}[htbp]
\centering
\fbox{\begin{minipage}{0.94\linewidth}
\small
\textbf{Explanation $e$ of donor site $t=1851$} (the site before ``Wait, that's
a problem.'' in Figure~\ref{construction:trace}).
\medskip

\textit{Paragraph 1 (kept):} The text establishes a formal, academic register
regarding a math problem, specifically highlighting a contradiction in the
solution process.
\smallskip

\textit{Paragraph 2 (kept):} The sentence structure ``But this seems to be
wrong because the solution is inconsistent with the assumption, so I must have
made a mistake in my thinking.'' creates a strong semantic expectation for a
concluding clause or elaboration on the error.
\smallskip

\textit{Paragraph 3 (kept):} The final token ``is inconsistent with the
assumption.''
\smallskip

\textit{Paragraph 4, original (prediction):} This indicates the immediate next
tokens must complete the thought, likely with ``Or perhaps I made a
miscalculation?'' or ``So why is this happening?'' to resolve the confusion.
\smallskip

\textit{Paragraph 4 in $T(e)$ (continue edit):} This ends the sentence,
implying the next token likely starts a new sentence or clause, such as
``Therefore, the total is calculated as follows.'' to continue the derivation.
\end{minipage}}
\caption{A real AV explanation and its edited version $T(e)$. The edit replaces
only the last paragraph. The text is verbatim. It comes from a later rebuild of
the direction, $N'$, which has cosine 0.996 with the deployed $N$. The
explanations behind the deployed $N$ were not saved, so they may differ slightly
from this text. The AV gets the kind of moment right: the model has just found a
contradiction. But it also invents details: the sentence quoted in paragraph 2
is not in the trace, although its meaning is close.}
\label{fig:explanation}
\end{figure}

The AR maps both texts back to vectors. Their difference $\delta_i$ is the
change in the state that the edit asks for. We scale each $\delta_i$ to unit
length, average the unit vectors over all $K$ donor sites ($K=35$ for our
donor trace), and scale the mean to unit length again:
\begin{equation}
 \delta_i=\mathrm{AR}(T(e_i))-\mathrm{AR}(e_i),\qquad
 N=\frac{\sum_{i=1}^{K}\delta_i/\lVert\delta_i\rVert_2}
 {\left\lVert\sum_{i=1}^{K}\delta_i/\lVert\delta_i\rVert_2\right\rVert_2}.
 \label{eq:edit}
\end{equation}
We do not normalize the two AR outputs before we take their difference. Section~\ref{inference:local} also tests single edits without averaging: at
each test boundary, it adds that boundary's own unit difference
$\delta_i/\lVert\delta_i\rVert_2$ once, at that boundary only.

One donor trace was enough in our tests. Directions built from more traces or
more sites shorten outputs about as much as $N$. Their pairwise cosines are
0.93--0.999, and none clearly works better (Appendix~\ref{construction:appendix}).
``One trace'' refers only to building $N$: the NLA itself needs the training
run of Section~\ref{method:interface}. We did not test whether the AV's
activation-conditioned explanation is needed (Section~\ref{sec:discussion}).

\subsection{The steering policy}
\label{method:policy}
During generation, VERS changes the state $h_t$ at a layer about 70\% of the
way through the model (layer 20 for the 1.7B; Section~\ref{sec:transfer} gives
the layers of the larger models) at each eligible token $t$:
\begin{equation}
 h_t'=h_t+\alpha\lVert h_t\rVert_2N.
 \label{eq:steer}
\end{equation}
$N$ has unit norm, so the added vector has norm $\alpha\lVert h_t\rVert_2$. At
$\alpha=1$, the added vector is as long as the current state itself. The random-direction controls in Section~\ref{inference:frozen} add
random unit vectors with the same $\alpha$ and schedule.

A token is \emph{eligible} if it is a paragraph boundary inside thinking and its
generated index is at least the \emph{delay} $d$. The index counts generated
tokens from one. The thinking opener \texttt{<think>} has index 1, and prompt
tokens are not counted. The policy for the 1.7B model uses $\alpha=1$ and
$d=256$. It does not detect answers, does not predict whether a check is useful,
and does not look at the next paragraph. For larger models
(Section~\ref{sec:transfer}), we map $N$ into the target model, normalize it
again, and choose $\alpha$ on the target's development data.

Appendix~\ref{sec:evidence} gives the implementation details.

\section{Controlling Reasoning at Inference Time}
\label{sec:inference}
This section tests VERS on the 1.7B source model. We first describe the
evaluation setup. Then we test the local edit behind the direction, evaluate
the frozen policy, and compare it with other methods.

\subsection{Evaluation and policy selection}
\label{sec:setup}
All models are from the Qwen3 family \cite{qwen2025qwen3}. The source model is
Qwen3-1.7B in thinking mode. Questions come from GSM8K \cite{cobbe2021gsm8k},
MATH \cite{hendrycks2021math}, and four math competitions: AIME 2024, AIME 2025, AMC 2023, and HMMT
February 2025. We give each question
set (\emph{cohort}) a short name. There are four main test cohorts.
\emph{gsm997} holds the 997 GSM8K test questions not used in earlier
experiments. The model solved each of them in 4 of 4 earlier samples (base
accuracy 98.92\%), so gsm997 is an easy subset, not natural GSM8K
(Appendix~\ref{appendix:cohorts}). Section~\ref{inference:frozen} also reports
the full GSM8K test (gsm1319).
\emph{MATH-500} is the standard 500-question subset of the MATH test set. \emph{confirm400} has 400 other
MATH test questions with the same level mix as MATH-500. \emph{hard128} has 128
competition problems. gsm997,
MATH-500, and confirm400 share no question text with each other or with the
development questions. Appendix~\ref{appendix:cohorts} lists every cohort, with its source,
size, number of samples, and use.

\paragraph{Sampling and scoring.}
Whole generations use temperature 0.6, top-$p$ 0.95, top-$k$ 20, and one shared
instruction to give the final answer in \texttt{\textbackslash boxed\{\}}. An
answer is correct if it is symbolically equal to the reference
answer. Some MATH reference answers are written in short LaTeX without
braces, for example \texttt{\textbackslash frac43}. Before grading, we rewrite
these into standard form, such as \texttt{\textbackslash frac\{4\}\{3\}}; this
changes 8 MATH-500 and 8 confirm400 references.
Outputs with no answer, or that reach the token budget, count as wrong. We
report the mean number of generated tokens, including the final answer. When a
question has several samples, we first average over them. 95\% intervals for
paired differences come from a bootstrap over questions. The one exception is
accuracy on gsm997, which uses a normal approximation to the paired
question-level differences.

\paragraph{Accuracy tolerance.}
Before testing, we fixed an accuracy tolerance of $-2$ percentage points (pp).
A method is \emph{non-inferior} to a reference if the lower end of the 95\%
interval for its accuracy change is above $-2$ pp. Non-inferiority allows a
small loss; it does not mean ``no loss''.

\paragraph{Directions and policy selection.}
We compare $N$ with $D$, a direction built without the NLA. To build $D$, we
take 297 correct 1.7B traces (dpairs297): 244 from GSM8K, 38 from AMC 2023,
and 15 from AIME 2025. In each trace, we read the state at two boundaries of
similar depth: one right before a ``Wait'' paragraph and one before an ordinary
paragraph. $D$ is the mean of the ordinary states minus the mean of the ``Wait''
states, scaled to unit length.

Each direction needs a strength $\alpha$ and a delay $d$. We chose them on
development questions that share no text with any test set: first a screen on
200 questions (dev200, one sample each), then a choice on 400 questions (dev400,
two samples each). These are mostly GSM8K questions taken from all difficulty
levels, so the base model solves only about 55\% of them. We took the setting with the fewest tokens
whose accuracy estimate stayed within the $-2$ pp tolerance set above. This gave $\alpha=1, d=256$ for $N$ and $\alpha=0.5, d=512$ for $D$.
We then froze both settings. All test runs use four samples per question.
Appendix~\ref{inference:delay} gives the full selection and a delay ablation.

\subsection{Do explanation edits change the next reasoning step?}
\label{inference:local}
Before we average many edits into $N$, we check that a single edit does what
its text says. We take 200 correct traces (local200). In each trace, we pick two
boundaries: one before an ordinary paragraph, where the model did not doubt, and
one right before a ``Wait'' paragraph, where it did. At each boundary we apply
one edit: the AV explains the state, we edit the explanation, and we add the
unit difference of the two AR outputs at that boundary only, once, with a small
strength $\alpha=0.25$. Then the model writes 60 more tokens, and we count how
often its next paragraph \emph{starts with doubt}, that is, has a doubt marker
(Section~\ref{construction:evidence}) in its first sentence. As a control, we
add a random vector of the same length instead of the edit.

\begin{table}[htbp]
\centering\small
\caption{One edit at one boundary ($\alpha=0.25$). Each value is the share of
60-token continuations whose next paragraph starts with doubt. 200 boundaries
of each kind; 4 samples per ordinary boundary and 8 per boundary before
``Wait''.}
\label{inference:localtable}
\begin{tabular}{lrr}
\toprule
Added at the boundary & Before ordinary paragraph & Before ``Wait''\\
\midrule
Nothing (base) & 8.8\% & 96.5\%\\
Doubt edit & 35.1\% & 96.7\%\\
Continue edit & 6.5\% & 74.2\%\\
Random vector, same length & 6.5\% & 94.1\%\\
\bottomrule
\end{tabular}
\end{table}

Each edit pushes the next step the way its text says
(Table~\ref{inference:localtable}). At ordinary boundaries the model rarely
doubts (8.8\%). The doubt edit raises this to 35.1\%, while the continue edit
and the random vector do not. Right before ``Wait'' the model almost always
doubts (96.5\%). The continue edit lowers this to 74.2\%, while the doubt edit
and the random vector leave it near base. We use a small strength on purpose:
at larger strengths even the control edits suppress doubt, so the test could no
longer show that the effect follows the meaning of the edit. This test covers
one edit at one boundary. The rest of this section tests the averaged direction
$N$ over whole generations.

\subsection{Behavior of the frozen policy}
\label{inference:frozen}
We compare the frozen VERS policy with three baselines:
\begin{enumerate}
\item $D$, the difference-of-means direction (Section~\ref{sec:setup});
\item a \emph{concision prompt}, with no steering: the system prompt adds
``Solve concisely. Avoid repeating calculations or explanations.'';
\item three random unit directions, with the same $\alpha$ and schedule as VERS.
\end{enumerate}
One fixed direction built from edited explanations shortens output by 45.8\% on
the full GSM8K test, with an accuracy change of $-0.82$ pp
(Table~\ref{inference:gsmfulltable}). On gsm997 it saves 42.9\% at $-0.78$ pp,
and on MATH-500 it saves 31.3\% at $-3.05$ pp (Table~\ref{inference:frozentable}).

VERS is non-inferior on gsm997: the lower end of its interval, $-1.29$ pp, is
above $-2$ pp. The loss is small but real, because the interval excludes zero.
On MATH-500, no method is non-inferior. $D$ and the prompt save fewer tokens and
also lose less accuracy. Section~\ref{sec:limits} shows where the losses
happen. Random vectors with the same strength do not shorten output, so adding
a large vector does not by itself explain the savings. Repeated steering also
does not wear off, and its edits do not build up in long answers. Edits of the
hidden state stay in the attention cache, and edits at every token can make the
output incoherent
\cite{kang2026promptactivationdualityimprovingactivation}. In an earlier test
with steering at every paragraph break, detected doubt stayed 99--100\% removed
from the first to the 50th paragraph. When we ran 789 steered answers again with
and without the hook, the edits in the cache barely changed later tokens: 99.1\%
of their probabilities changed by less than 20\% (Appendix~\ref{appendix:longtrace}). The concision prompt is
a strong simple baseline. On gsm997 it saves 37.0\% at $-0.55$ pp, and VERS
saves 42.9\% at $-0.78$ pp. So on easy questions, VERS gains little over a
prompt.

\paragraph{Full GSM8K test.} gsm997 holds only questions that the model
already solves. To test natural GSM8K, we also ran base, VERS, and the
concision prompt on all 1,319 GSM8K test questions (gsm1319, four samples each)
in one run (Table~\ref{inference:gsmfulltable}). VERS saves 45.8\% of tokens at
$-0.82$ pp, and it is still non-inferior. On the gsm997 part, the run repeats
the result above. On the other 322 questions, where base solves only 63.3\%,
VERS saves 51.0\% with no clear accuracy change. Here the concision prompt
raises accuracy by 2.64 pp, and VERS is 2.80 pp $[-5.43,{-}0.16]$ below the
prompt. So on harder questions, the prompt can be the better choice.

\begin{table}[htbp]
\centering\small
\caption{The full GSM8K test (gsm1319, four samples per question, one run).
Each method is compared with base from the same run. Accuracy changes are in
pp with 95\% bootstrap intervals over questions.}
\label{inference:gsmfulltable}
\begin{tabular}{lrrrrr}
\toprule
& & \multicolumn{2}{c}{VERS ($N$)} & \multicolumn{2}{c}{Concision prompt}\\
\cmidrule(lr){3-4}\cmidrule(lr){5-6}
Questions & Base acc. & Tokens saved & $\Delta$ accuracy & Tokens saved & $\Delta$ accuracy\\
\midrule
\rowcolor{black!10}All 1,319 & 90.3\% & \textbf{45.8\%} & $-0.82\ [\mathbf{-1.42},{-}0.21]$ & 34.9\% & $-0.09\ [-0.83,{+}0.63]$\\
gsm997 part & 99.1\% & 42.6\% & $-1.03\ [-1.50,{-}0.58]$ & 36.6\% & $-0.98\ [-1.55,{-}0.45]$\\
Other 322 & 63.3\% & 51.0\% & $-0.16\ [-2.25,{+}1.86]$ & 32.1\% & $+2.64\ [{+}0.08,{+}5.36]$\\
\bottomrule
\end{tabular}
\end{table}

Removing doubt markers is not the same as shortening. Under the frozen
policies, $D$ removes fewer detected doubt paragraphs than VERS (Table~\ref{inference:frozentable}),
so the two cannot be compared on this point. A separate, earlier test does
compare them at the same level of doubt removal. On 150 questions (dom150: 149
GSM8K questions that the model did not always solve, and one AIME 2025
question), we steered with each direction at $\alpha=1$ at every paragraph
boundary, with one sample per question. Both removed about 90\% of detected
doubt paragraphs (VERS 90\%, $D$ 91\%). But VERS shortened median thinking by
55\%, and $D$ by only 27\%. So VERS does more than remove doubt markers. The
accuracy changes in this small test are uncertain: $-4.7$ pp
$[-13.8,{+}4.5]$ for VERS and $-2.7$ pp $[-10.3,{+}5.0]$ for $D$.

\begin{table}[H]
\centering\small
\setlength{\tabcolsep}{4.5pt}
\caption{Frozen-policy results. VERS rows are shaded and the main results are
in bold. VERS saves the most tokens on both
datasets (42.9\% and 31.3\%) and removes the most detected doubt. On gsm997 its
loss stays within the $-2$ pp tolerance (lower end $-1.29$, bold). On MATH-500
the saving costs 3.05 pp, more than $D$ or the prompt lose. Random vectors with
the same strength make outputs longer, not shorter. Mean length counts all generated tokens
(thinking and final answer). ``Detected doubt reduced'': share of doubt
paragraphs (English marker rule, Section~\ref{construction:evidence}) that
disappear relative to base. Accuracy changes have 95\% intervals for the paired
per-question difference: from a normal approximation on gsm997, and from
resampling questions (bootstrap) on MATH-500 (Section~\ref{sec:setup}). Each dataset is
compared with its own base. ``Random'' adds three random unit vectors, one at a
time, with the same $\alpha$ and schedule as VERS. Its row gives the range over
the three vectors, which have no intervals. Negative values mean longer output
or more detected doubt.}
\label{inference:frozentable}
\begin{tabular}{llrrrr}
\toprule
Dataset & Method & Mean length & Tokens saved & \begin{tabular}[b]{@{}r@{}}Detected doubt\\reduced\end{tabular} & $\Delta$ accuracy [CI], pp\\
\midrule
gsm997 & Base (98.92\%) & 1,813 & \multicolumn{3}{c}{reference}\\
\rowcolor{black!10}& VERS ($N$) & 1,035 & \textbf{42.9\%} & \textbf{90.4\%} & $-0.78\ [\mathbf{-1.29},{-}0.27]$\\
& $D$ & 1,327 & 26.8\% & 66.5\% & $-0.30\ [-0.72,{+}0.12]$\\
& Concision prompt & 1,143 & 37.0\% & 52\% & $-0.55\ [-1.10,{-}0.01]$\\
& Random (3 vectors) & 1,813--2,156 & $-18.9$ to 0.0\% & $-25$ to 18\% & $-0.2$ to $+0.1$\\
\midrule
MATH-500 & Base (91.15\%) & 5,602 & \multicolumn{3}{c}{reference}\\
\rowcolor{black!10}& VERS ($N$) & 3,848 & \textbf{31.3\%} & \textbf{88.9\%} & $-3.05\ [-4.40,{-}1.70]$\\
& $D$ & 4,477 & 20.1\% & 77.3\% & $-1.45\ [-2.60,{-}0.35]$\\
& Concision prompt & 4,263 & 23.9\% & 29\% & $-1.75\ [-3.05,{-}0.45]$\\
& Random (3 vectors) & 5,896--6,030 & $-7.6$ to $-5.2$\% & $-5$ to 33\% & $0.00$ to $+0.35$\\
\bottomrule
\end{tabular}
\end{table}

\subsection{Comparison with other interventions}
\label{inference:comparators}
We compare VERS with three published methods on confirm400, with four samples
per question:
\begin{itemize}
\item A NoWait-style \emph{word ban} \cite{wang2025nowait} blocks the tokens that
open doubt paragraphs. The \emph{matched ban} blocks them only at VERS's sites.
The \emph{global ban} blocks them everywhere.
\item SEAL \cite{chen2025seal} builds its direction from states in execution
paragraphs versus reflection and transition paragraphs.
\item DEER \cite{yang2025deer} tries a short answer at each ``Wait'' and stops
thinking when its confidence is above 0.95. We count its trial-answer tokens
as cost.
\end{itemize}
This comparison uses the same confirm400 questions as the transfer test
(Section~\ref{sec:transfer}), but in a separate run. In that run we evaluate the
base model again, with the same settings as the other methods, and compare every
method with this base. The two runs give slightly different results on the same
questions (base accuracy 91.4\% here and 90.2\% in Table~\ref{transfer:tab-behavior}). Appendix~\ref{appendix:comparators} gives
all settings, intervals, and results.

\begin{figure}[htbp]\centering
\includegraphics[width=0.62\linewidth]{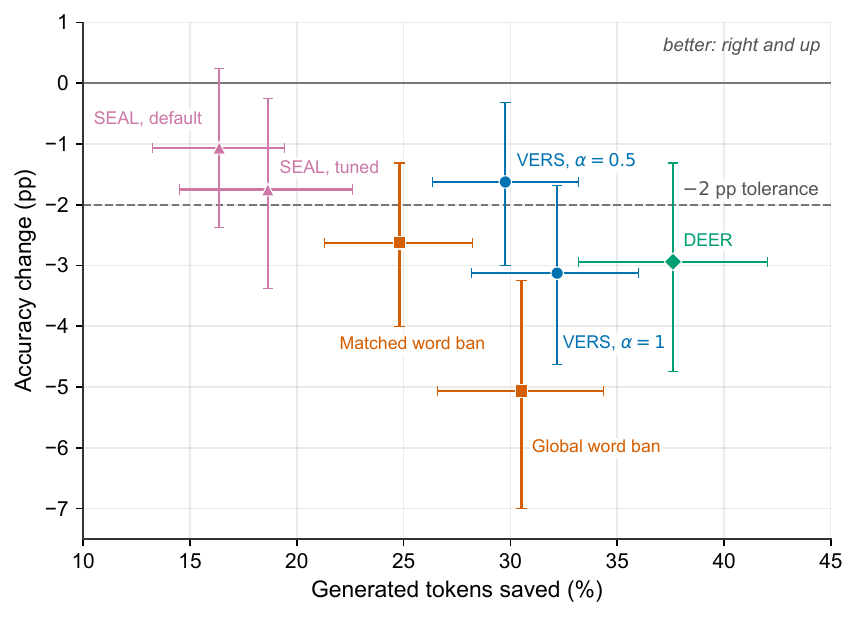}
\caption{Tokens saved and accuracy change of each method compared with base, on
confirm400 (Qwen3-1.7B, 400 questions, four samples each). Up and to the right
is better. Bars are 95\% paired bootstrap intervals over questions, in both
directions. The dashed line is the $-2$ pp tolerance. VERS uses delay 256.
DEER's cost includes its trial-answer tokens. These intervals compare each
method with base; the pairwise comparisons in the text use
Table~\ref{inference:pairwisetable}.}
\label{fig:baseline-comparison}
\end{figure}

Against base (Figure~\ref{fig:baseline-comparison}), full-strength VERS saves
32.2\% at $-3.12$ pp $[-4.62,{-}1.69]$. Half-strength
VERS saves 29.8\% at $-1.62$ pp $[-3.00,{-}0.31]$, so it is not non-inferior to
base. The matched ban saves 24.8\% at $-2.62$ pp, and tuned SEAL saves 18.6\%
at $-1.75$ pp. The global ban saves 30.5\% but loses 5.06 pp. DEER saves the
most, 37.6\%, at $-2.94$ pp.

We also compare methods directly (Table~\ref{inference:pairwisetable}). Before
any run, the comparator protocol listed $\alpha=0.5$ as a second VERS setting
and fixed a pairwise rule: a method passes against another if it saves more
(interval above zero) and the accuracy difference has a lower end of at least
$-2$ pp. These intervals are not adjusted for multiple comparisons.
Half-strength VERS passes against the matched ban (accuracy difference $+1.00$
pp $[-0.38,{+}2.38]$) and against tuned SEAL ($+0.12$ pp $[-1.31,{+}1.62]$).
SEAL's unit vector on VERS's schedule at the same strength saves 19.5\%, with a
difference of $+0.62$ pp $[-0.81,{+}2.06]$. Full-strength VERS saves more than the matched ban, but does not pass the
accuracy part of the rule. DEER passes against full-strength VERS: its
accuracy minus that of VERS is $+0.19$ pp $[-1.62,{+}1.88]$. VERS changes the
continuation, while DEER mainly stops it.

\paragraph{Is reasoning after the answer useful?}
Early-exit methods such as DEER stop thinking once the answer looks settled. We
test what stopping costs, and whether VERS can shorten the same reasoning
without that cost. On 387 development questions (cont387), the model thinks
until its answer settles: at each paragraph break we ask it for an answer, and
we stop when four answers in a row agree. From exactly this text, we then
continue in four ways: (1) return the settled answer; (2) close thinking and let
the model write an answer; (3) keep thinking normally; (4) keep thinking with
VERS. Compared with thinking normally, returning the settled answer loses
13.7 pp, and closing thinking right away loses 3.9 pp. So the reasoning after a
settled answer often still fixes it. VERS cuts this later thinking by 75\%, with
an accuracy change of $-0.3$ pp $[-3.2,{+}2.7]$. This interval is too wide to
show equal accuracy, but VERS shortens the later reasoning instead of stopping
it. Appendix~\ref{appendix:continuation} gives the full setup and results, and
Appendix~\ref{appendix:ellipse} shows one solution shortened by each method.

\section{Transferring the Intervention to Larger Models}
\label{sec:transfer}

Training the NLA on the 1.7B source model is expensive, so we test whether
parts of it can be reused on larger Qwen3 models. We make two claims of
different strength. (a) \emph{Direction transfer:} the mapped steering
direction $N$ cuts tokens and detected doubt paragraphs at every tested size,
up to 235B-A22B (Section~\ref{transfer:results}). (b) \emph{Interface
transfer:} reading the larger models with the 1.7B NLA works only partly
(Section~\ref{transfer:fidelity}). Through linear maps, the unchanged 1.7B AV
can explain 8B and 30B-A3B states, and the 1.7B AR can rebuild them. But the
rebuilt states keep much less of the original (FVE 0.334--0.386, against 0.632
on the 1.7B itself). The explanations also carry only a small signal that is
specific to the larger model. Both depend on fitted affine maps. Steering also
uses a strength $\alpha$ tuned for each model on development data. All models are from
one family, Qwen3.

\subsection{Mapping the direction into larger models}
\label{transfer:method}

Following affine model alignment~\cite{chen2025stitching}, we fit two ridge
maps. Here $h_s$ is a source state and $h_t$ is a target state at the same
token. $E(h_t)=W_Eh_t+b_E$ maps a target state into the source space, so the
frozen AV can read it. $U(h_s)=W_Uh_s+b_U$ maps a source-space vector, such as
an AR reconstruction or $N$, back into the target space. We fit the maps on
pairs $(h_s,h_t)$ from the same text run through both models; Qwen3 models
share one tokenizer. The target layer keeps the source layer's relative depth
(column $l/L$ of Table~\ref{transfer:tab-behavior}). We follow the original
NLA work and use a layer at about 70\% of the depth, and we did not tune it. A
small sweep on the 8B (layers 18 to 30) found no clearly better layer for our
task: layers 22 to 30 behaved similarly, and layer 18 was too weak. We do not
claim that this choice is optimal. Appendix~\ref{appendix:transfer} gives the
details.

Steering uses $N_t=W_UN/\|W_UN\|_2$. The policy is the source policy
(Eq.~\ref{eq:steer}): steer at paragraph boundaries after generated token 256,
with strength scaled by the current target residual norm. We choose $\alpha_t$
for each target on its own development data. We use a fixed \emph{development accuracy rule}: a
strength qualifies if, on development data, accuracy falls by at most 1 pp and
the share of outputs that reach the token limit at most doubles. Among the
strengths that qualify, we take the one whose token saving is closest to the
1.7B's development saving (48.8\%). Steering does not use the AV or
the AR.

\subsection{The steering direction transfers}
\label{transfer:results}

With fitted maps and a tuned $\alpha$, the mapped direction transfers to all
three larger models. On confirm400, it saves 30--35\% of tokens on every target
and meets every pre-stated criterion (Table~\ref{transfer:tab-behavior}). The
token saving must be within five points of the source's saving. Detected doubt
paragraphs must fall by at least 85\%. The accuracy change must be at most 1 pp
worse than the source's change.

\begin{table}[htbp]
\centering\small
\begin{adjustbox}{max width=\linewidth}
\begin{tabular}{lrrrrll}
\toprule
Model & $l/L$ & $\alpha$ & Base acc. & Detected doubt reduced & $\Delta$ accuracy, pp [95\% CI] & Tokens saved [95\% CI]\\
\midrule
1.7B & 20/28 & 1.00 & 90.2\% & 89.4\% & $-2.1$ [$-3.7$, $-0.4$] & 30.2\% [25.8, 34.4]\\
8B & 26/36 & 0.75 & 97.8\% & 92.0\% & $-0.1$ [$-0.9$, $+0.6$] & 31.9\% [28.9, 34.9]\\
30B-A3B & 34/48 & 0.75 & 99.0\% & 87.9\% & $-0.6$ [$-1.1$, $-0.1$] & 30.2\% [26.8, 33.5]\\
235B-A22B & 67/94 & 1.50 & 99.2\% & 91.3\% & $-0.8$ [$-1.4$, $-0.2$] & 35.1\% [32.2, 37.8]\\
\bottomrule
\end{tabular}
\end{adjustbox}
\caption{Steering transfer on confirm400 (four samples per question). Tokens
saved is the drop in mean generated tokens (thinking and final answer) relative
to the base model of the same size; $\Delta$ accuracy is also relative to that
base. Detected doubt reduced is the drop in detected doubt paragraphs. $\alpha$ is tuned for
each model on development data. The 1.7B row is a separate run from the
comparator run in Appendix~\ref{appendix:comparators}, with its own base. The
235B uses FP8 weights and bf16 activations.}
\label{transfer:tab-behavior}
\end{table}

Confirm400 cannot show that the larger models keep their reasoning quality:
their base accuracy is near the ceiling (97.8--99.2\%). On the hard competition
set hard128, every size loses accuracy at its primary strength: $-7.0$ pp
(1.7B), $-3.2$ pp (8B), $-5.4$ pp (30B-A3B), and $-5.6$ pp (235B-A22B)
(Section~\ref{limits:hard}).

Three controls point to $N$, not the maps, as the source of the saving
(Table~\ref{transfer:tab-controls}). Random source directions sent through the
same maps make output longer. A difference-of-means direction $D$ built from
each larger model's own states, and tuned separately, saves less than mapped
$N$. On the 8B, the source $D$ sent through the same map saves less still, so
the gap between $N$ and $D$ is not due to the alignment alone.

\begin{table}[htbp]
\centering\small
\begin{tabular}{lrrr}
\toprule
Direction added in the larger model & 8B & 30B-A3B & 235B-A22B\\
\midrule
VERS: 1.7B direction $N$, mapped & 31.9\% & 30.2\% & 35.1\%\\
Difference of means $D$, built in the larger model & 26.7\% & 25.9\% & not run$^\ddagger$\\
Difference of means $D$ from the 1.7B, mapped & 22.0\% & 30.6\%$^\dagger$ & not run$^\ddagger$\\
Random 1.7B direction, mapped & $-6.6$\% & $-7.2$\% & $-3.5$\%$^\dagger$\\
\bottomrule
\end{tabular}
\caption{Does the saving in larger models come from $N$, or from the maps?
Only $N$ saves this much. A random direction sent through the same maps makes
outputs longer, and $D$ saves less, whether it is built in the larger model or
mapped from the 1.7B. Values are tokens saved relative to each model's base on
confirm400; negative values mean longer output. $^\dagger$On development data, not confirm400. On the same 30B-A3B development
questions, mapped $N$ saves 42.4\% at $-0.2$ pp, while the mapped 1.7B $D$ saves
30.6\% at $-1.9$ pp.
$^\ddagger$The 235B-A22B test covered only $N$ and a random control.}
\label{transfer:tab-controls}
\end{table}

\subsection{How much of the interface transfers?}
\label{transfer:fidelity}

Steering needs only the mapped direction $N$. Here we ask a separate question:
can the unchanged 1.7B NLA also read the larger models? We test two things on
the 8B and the 30B-A3B; we did not test the 235B-A22B.

\paragraph{Rebuilding states.} We send a state of the larger model through the
maps, the 1.7B AV, and the 1.7B AR, and compare the result with the true state
(FVE, Eq.~\ref{eq:fve}). On 10,000 fresh web and chat states, FVE is 0.334 on
the 8B and 0.386 on the 30B-A3B, against 0.632 for the 1.7B NLA on its own
states (Table~\ref{transfer:tab-fidelity} in Appendix~\ref{appendix:transfer}).
Most of the loss comes from the text step: the same maps without the AV and AR
reach 0.584 and 0.686.

\paragraph{Reading the larger model.} A useful reader should describe what the
larger model will do, not only the text it has read. At about 964 paragraph
boundaries per model, the 1.7B and the larger model read the same text but
differ in how likely they are to start doubting next. We test whether the
explanations of the two models' states differ in the same direction as this
gap. They do, but only weakly: Spearman $\rho$ is $+0.09$ to $+0.19$ on the 8B
and $+0.07$ to $+0.12$ on the 30B-A3B, over three ways of scoring doubt. So the
explanations mostly describe the text, and only a little of what is specific
to the larger model. Appendix~\ref{appendix:transfer} gives the details.

\section{Where Control Fails}
\label{sec:limits}
\subsection{Hard problems keep an accuracy cost across scale}
\label{limits:hard}
On hard competition problems, every method we tested has a lower accuracy
estimate than base, at every model size (Table~\ref{limits:hardtable}). We test
on hard128 (128 problems from AIME 2024 and 2025, AMC 2023, and HMMT February
2025), with eight seeds per problem. It leaves out two AIME 2025 problems:
the donor-trace problem and one development problem.

\begin{table}[htbp]
\centering\small
\textbf{(a) VERS at every model size}\\[2pt]
\begin{tabular}{lrrrr}
\toprule
& 1.7B & 8B & 30B-A3B & 235B-A22B\\
\midrule
Strength $\alpha$ & 1.0 & 0.75 & 0.75 & 1.5\\
Base accuracy (\%) & 50.0 & 72.7 & 76.6 & 83.4\\
$\Delta$acc (pp) & $-7.0$ & $-3.2$ & $-5.4$ & $-5.6$\\
Tokens saved (\%) & 14.1 & 11.1 & 11.6 & 22.5\\
\bottomrule
\end{tabular}

\vspace{8pt}
\textbf{(b) All methods on the 1.7B and the 8B}\\[2pt]
\begin{tabular}{lrrrr}
\toprule
& \multicolumn{2}{c}{1.7B} & \multicolumn{2}{c}{8B}\\
\cmidrule(lr){2-3}\cmidrule(lr){4-5}
Method & $\Delta$acc & TS & $\Delta$acc & TS\\
\midrule
VERS, main strength & $-7.0$ & 14.1 & $-3.2$ & 11.1\\
VERS, $\alpha=0.5$ & $-5.6$ & 11.7 & \multicolumn{2}{c}{not run}\\
$D$, built in the same model & $-5.7$ & 9.3 & $-2.4$\rlap{$^{\dagger}$} & 8.8\\
$D$ from the 1.7B, mapped & \multicolumn{2}{c}{same as row above} & $-2.7$ & 8.7\\
Matched word ban & $-6.5$ & 12.6 & $-2.6$ & 13.1\\
SEAL, default & $-4.3$ & 3.7 & $-2.1$\rlap{$^{\dagger}$} & 6.3\\
DEER & $-2.2$\rlap{$^{\dagger}$} & 7.6 & $-2.7$ & 8.5\\
\bottomrule
\end{tabular}
\caption{Hard128 results. $\Delta$acc: accuracy change from the same model's
base (pp); a negative value means lower accuracy than base. TS (tokens saved):
drop in mean generated tokens relative to the same base (\%); for DEER this
counts its trial tokens. Panel (a) covers every size; the other methods were run
only on the 1.7B and the 8B (panel b). On the 30B-A3B, $D$ built in the same
model gives $-3.9$ pp and saves 5.0\%. $\dagger$: the 95\% accuracy interval
includes zero. No configuration meets the $-2$ pp non-inferiority margin.}
\label{limits:hardtable}
\end{table}

No configuration is non-inferior: for every one, the lower end of the accuracy
interval is below our margin of $-2$ pp. For three configurations (DEER on the
1.7B, native $D$ and SEAL on the 8B), the interval includes zero, so their loss
is not certain. Halving the 1.7B strength does not clearly bring accuracy back:
VERS loses 5.6 pp at $\alpha=0.5$ and 7.0 pp at $\alpha=1.0$. At $\alpha=1.0$, the 1.7B
also switches into Chinese in 37\% of rollouts. Too much strength
also causes loops: at $\alpha=2$, the 235B reaches the token limit in 39\% of
rollouts.

Larger models do not reliably remove the cost (Figure~\ref{fig:competition-scale} in
Appendix~\ref{appendix:selectivity}). Compared with the 1.7B, which loses
7.0 pp, the 8B loses 3.8 pp less (95\% interval $[+0.4,{+}7.1]$). For the
30B-A3B and the 235B-A22B the difference is uncertain: 1.7 pp less
$[-2.9,{+}6.0]$ and 1.5 pp less $[-2.5,{+}5.5]$. Both intervals include zero,
so we cannot say that these models lose less than the 1.7B.

\paragraph{Harder problems need more of their checks.}
On MATH-500, the accuracy cost grows with difficulty. The VERS accuracy changes
for levels 1 to 5 are 0.0, $-0.8$, $-1.0$, $-2.0$, and $-8.2$ pp (level 5:
[$-12.13$, $-4.29$]), and level 5 also has the smallest token saving, 24.8\%.
The larger losses on hard problems are consistent with steering away checks
that do useful work. But our tests do not separate this cause from other
effects of steering. If it is the cause, a better policy would steer only the
checks that are not needed. Section~\ref{limits:selectivity} tests two ways to do this.

\paragraph{Trace quality.}
This audit checks that VERS traces still read as clear solutions and not as
broken text. It does not measure the overall quality of the reasoning. A blind
judge (gpt-5.6-luna, low reasoning effort) read 200 pairs of correct traces from
MATH-500, 40 per level, one pair per question where both base and VERS were
correct. It saw one solution at a time, with the problem and the reference
answer, but not which method wrote it. For each solution it answered whether it
would ``accept this solution as a worked example to teach a student''.\footnote{During this audit, the OpenAI account we used was suspended
for policy violations. We suspect that this wording was treated as an attempt
at distillation, and we recommend not using it as is.}
We call the share of ``yes'' answers \emph{acceptance}
(Table~\ref{limits:qualitytable}).

\begin{table}[htbp]
\centering\small
\begin{tabular}{lrrr}
\toprule
Traces judged & Base & VERS & Difference (pp)\\
\midrule
All levels (200 pairs) & 85.0\% & 83.5\% & $-1.5$ $[-6.5,{+}3.5]$\\
Level 5 (40 pairs) & 75.0\% & 67.5\% & $-7.5$\\
\bottomrule
\end{tabular}
\caption{Share of correct MATH-500 traces that a blind judge accepts as a worked
example. The 95\% interval is given for all levels together.}
\label{limits:qualitytable}
\end{table}

Acceptance is close for base and VERS, so the audit gives no clear sign that
VERS traces are worse. The small drop, larger at level 5, may come from steering
at every paragraph break, which can break the flow of the argument from one
paragraph to the next. We did not test this.

\subsection{Two selective constructions do not improve the trade-off}
\label{limits:selectivity}
We tried two ways to steer more selectively, so that the few doubt episodes
that fix an error stay unsteered. Neither improved the observed accuracy--length
trade-off. This does not show that selective control is impossible; we tested
only these two constructions, and we encourage future research in this
direction.

A \emph{doubt episode} starts at a doubt paragraph and runs for at most four
paragraphs. It is \emph{productive} if the model's forced answer changes from
wrong before it to right after it, and \emph{redundant} if the answer stays
right. These labels are proxies: ``redundant'' does not prove that the reasoning
was unnecessary. In probe1000 (1,000 base 1.7B rollouts on MATH training
questions), 1.8\% of doubt episodes are productive and 41\% are redundant; most
of the rest stay wrong (Appendix~\ref{appendix:selectivity}).

First, we check that the state holds the needed information. We fit two
logistic-regression probes on the layer-20 state. Probe A, read just before a
doubt paragraph, tells productive from redundant episodes with AUC 0.75. Probe
B, read at every paragraph boundary, predicts whether a productive episode
starts next, with AUC 0.84. So the state does carry some signal about which
checks are useful.

We then tried two ways to use this signal (Table~\ref{limits:seltable}):
\begin{enumerate}
\item \emph{Build the direction only from redundant doubt}, so that it does not
push away useful checks. This direction is almost the same as one built from
productive doubt or from all doubt (cosine 0.98--0.99), and it behaves like $N$.
\item \emph{Gate the steering with probe B}. During generation, when probe B
predicts a useful check, we skip steering at that boundary and the next three.
We compare the gate with random skipping that leaves the same share of
boundaries unsteered. The gate does not keep more accuracy, and it saves fewer
tokens.
\end{enumerate}

\begin{table}[htbp]
\centering\small
\begin{tabular}{llrr}
\toprule
Test set & Steering & $\Delta$acc (pp) & Tokens saved\\
\midrule
sel400 & $N$ & $-3.3$ $[-4.9,{-}1.6]$ & 29.5\%\\
& $N$ built only from redundant doubt & $-3.7$ $[-5.4,{-}2.0]$ & 30.2\%\\
\midrule
gate400 & Random skipping (control) & reference & 29.5\%\\
& Gate with probe B & $-1.0$ $[-2.7,{+}0.6]$ & 20.1\%\\
\bottomrule
\end{tabular}
\caption{Two selective constructions on hard MATH questions (sel400 and gate400:
400 level-4 and level-5 MATH test questions each, four samples per question).
Tokens saved is relative to base. $\Delta$acc is relative to base on sel400,
and relative to random skipping on gate400.}
\label{limits:seltable}
\end{table}

Appendix~\ref{appendix:selectivity} gives the label shares, the probes, the
gate, and the limits of these tests.

\subsection{Reversing \texorpdfstring{$N$}{N} adds doubt, but the doubt-edit direction does not}
\label{limits:reverse}
Can the steering also make the model doubt more? We tested two directions on
400 development questions (dev400, two samples each) at three strengths
(Table~\ref{limits:ptable}). The first is $-N$, the opposite of $N$. The second
is $P$, built in the same way as $N$: from the same 35 donor sites and the same
explanations, but with the doubt edit, which predicts that the next sentence is
``Wait, let me double-check that.''

\begin{table}[htbp]
\centering\small
\begin{tabular}{rrrrr}
\toprule
& \multicolumn{2}{c}{Doubt-edit direction $P$} & \multicolumn{2}{c}{Reversed direction $-N$}\\
\cmidrule(lr){2-3}\cmidrule(lr){4-5}
$\alpha$ & Longer output & $\Delta$ doubt paragraphs & Longer output & $\Delta$ doubt paragraphs\\
\midrule
0.25 & 3.4\% & $+0.17$ $[-1.75,{+}2.15]$ & 35.2\% & $+24.5$ $[{+}21.1,{+}28.4]$\\
0.5 & 10.3\% & $+1.26$ $[-0.65,{+}3.22]$ & 73.2\% & $+50.9$ $[{+}45.6,{+}56.9]$\\
1.0 & 14.7\% & $-1.13$ $[-2.97,{+}0.70]$ & 229.8\% & $+194.6$ $[{+}179.1,{+}212.2]$\\
\bottomrule
\end{tabular}
\caption{Steering toward doubt on dev400. $\Delta$ doubt paragraphs is the
change in detected doubt paragraphs per rollout, with 95\% intervals; base has
about 27--29. Each direction is compared with the base model from its own run.
$P$ makes outputs longer, but every interval for the change in doubt includes
zero. $-N$ adds doubt at every strength.}
\label{limits:ptable}
\end{table}

$-N$ adds doubt, and $P$ does not. At $\alpha=0.25$, $-N$ almost doubles the
detected doubt paragraphs (53.2 per rollout, against 28.7 for base), with an
accuracy change of $-0.62$ pp $[-3.50,{+}2.25]$. At $\alpha=1$ it breaks the
model: 55\% of outputs reach the token limit (base: 0.75\%), and accuracy falls
by 5.00 pp $[-8.00,{-}2.38]$ (Table~\ref{appendix:negntable}). So steering along $N$ works in both directions.
The failure of $P$ comes from the doubt edit, not from steering itself. By our
pre-set rule, $P$'s result ended its test, so we did not run $P$ on the hard
test sets.

Two things may explain why the doubt edit fails. First, the donor sites lie
right before doubt paragraphs, so the state may already hold most of what the
doubt edit asks for. The doubt edit also changes the AR output less than the
continue edit (Appendix~\ref{appendix:selectivity}). A single doubt edit at
ordinary boundaries did raise doubt from 8.8\% to 35.1\%
(Section~\ref{inference:local}), so the choice of donor sites may matter.
Second, $P$ is not the opposite of $N$: the two are almost orthogonal
($\cos(P,N)=0.022$). Both are differences from the same original explanation,
so nothing requires them to point in opposite directions.

\section{Learning from Steered Reasoning Traces}
\label{sec:distillation}
Can a model learn the shorter style from steered traces, and keep it without
the steering hook? We use Qwen3-8B. With the hook off, it writes natural
traces; with the VERS hook on (the \emph{hooked teacher}), it writes steered
traces. Both cover 6,268 MATH training questions. We then fine-tune four
Qwen3-8B students:
\begin{enumerate}
\item \emph{Short-SFT}: supervised fine-tuning on the shortest correct natural
trace of each question;
\item \emph{Steered-SFT}: the same, on the shortest correct steered trace;
\item \emph{DPO-natural} and \emph{DPO-steered} \cite{rafailov2023dpo}: DPO on
2,477 preference pairs each, built from natural or from steered traces.
\end{enumerate}
All students use rank-64 LoRA \cite{hu2021lora}, one epoch, and one training
seed, and they run without the hook. We did not fully fine-tune the 8B: on the
1.7B, LoRA and full fine-tuning gave the same verdict on every comparison of
steered with natural data
(Appendix~\ref{distill:small}). We test them on two MATH test sets that
share no questions with training: \emph{fresh1000} (1,000 questions, 200 per
level) and \emph{gate400} (400 level-4/5 questions), with four samples per
question. Appendix~\ref{distill:setup} gives the full setup.

\begin{table}[htbp]\centering\small
\begin{tabular}{lrrrr}\toprule
& \multicolumn{2}{c}{fresh1000 (mixed)} & \multicolumn{2}{c}{gate400 (hard)}\\
\cmidrule(lr){2-3}\cmidrule(lr){4-5}
Student & $\Delta$acc (pp) & Tokens saved & $\Delta$acc (pp) & Tokens saved\\\midrule
Short-SFT$^\dagger$ & $-0.52$ & 0.2\% & $-0.88$ & 2.1\%\\
\rowcolor{black!10}Steered-SFT & $-0.85$ & \textbf{33.2\%} & $-1.44$ & \textbf{28.7\%}\\
DPO-natural$^\dagger$ & $-0.12$ & 8.5\% & $-0.25$ & 7.7\%\\
DPO-steered$^\dagger$ & $-0.40$ & 9.7\% & $-0.50$ & 9.7\%\\\bottomrule
\end{tabular}
\caption{Hook-free 8B students, compared with the base model before student fine-tuning.
Shading marks the student with the largest token saving (Steered-SFT).
$^\dagger$Passes the $-2$ pp accuracy rule on both sets. Intervals and paired
comparisons are in Appendix~\ref{distill:setup}.}
\label{distill:tab-summary}
\end{table}

Steered-SFT learns most of the shorter style (Table~\ref{distill:tab-summary}):
without the hook, it saves 33.2\% of tokens on fresh1000 and 28.7\% on gate400.
It passes the $-2$ pp accuracy rule on fresh1000 but not on gate400. Training on
the model's own shortest natural traces does not shorten it (Short-SFT), so the
saving comes from the steered data. Both DPO students pass the $-2$ pp rule on both sets, but they
save much less (8--10\%), and steered data adds little to DPO. So steered
traces can teach a shorter style, but the cost on hard questions remains. The
1.7B students agree on all four points: Steered-SFT saves the most but loses
accuracy, Short-SFT barely shortens, DPO loses little accuracy, and steered data does
not improve DPO. The only difference is how much DPO saves: 8--10\% on the
8B, but 15--25\% on the 1.7B (Appendix~\ref{distill:small}).

\section{Discussion and Conclusion}
\label{sec:discussion}
\label{sec:conclusion}
We asked whether edits of activation explanations can steer how a model
reasons. Within the limits below, they can. One fixed direction, built from
explanation edits in a single trace, shortens reasoning on other math
questions. It transfers to larger Qwen3 models through affine maps, and
students fine-tuned on steered traces keep the shorter style without the hook.

\paragraph{In practice.} The accuracy cost is small on GSM8K ($-0.82$ pp on the
full test) and larger on MATH-500 and competition problems, so VERS fits best
where the base model already solves most questions. Even there, it gains
little over a concision prompt (Section~\ref{inference:frozen}), and among
published methods it is one option on the trade-off between tokens and
accuracy (Section~\ref{inference:comparators}). One trained NLA can serve
larger models through affine maps, which should be much cheaper than training
a new NLA, although we did not measure this cost. Through the maps, the NLA
rebuilds the larger models' states much less well (FVE 0.33--0.39, against
0.63 on the 1.7B). Steering only needs the mapped direction $N$, but work that
uses the NLA to read larger models should take this loss into account.

\paragraph{Limitations.}
All tasks are math problems, and all models come from the Qwen3 family. Of the
two edits we tested, only the continue edit gave a working control. We did not
test whether the AV is needed: the AR alone, given hand-written ``check'' and
``continue'' texts instead of the AV's explanations of real states, might give
a similar direction. A high reconstruction FVE does not by itself show that the
explanations are faithful. Our doubt detector counts English markers only
(Section~\ref{construction:evidence}). Each student was trained with one seed,
and token savings are not measured wall-clock speedups.

\paragraph{Future work.}
\begin{enumerate}
\item \emph{Test whether the AV is needed.} Build the direction with the AR
alone, from hand-written ``check'' and ``continue'' texts, and repeat our edits
without the kept context paragraphs of each explanation.
\item \emph{Train the NLA on reasoning.} Its FVE is 0.63 on web and chat text
but only 0.30 on reasoning states.
\item \emph{Adapt the NLA better to larger models}, for example by training an
interface for each model with RL, and compare it with the maps on
reconstruction, on a second edit, and on cost.
\item \emph{Find the checks that matter.} The main open problem is
selectivity: the direction does not tell the checks a model needs from the
ones it does not, so hard problems still lose accuracy. Better labels should
include necessary case analysis and error correction, and a gate built on them
must work during generation, where steering changes the later states
(Section~\ref{limits:selectivity}).
\end{enumerate}

\bibliographystyle{unsrtnat}
\bibliography{references}

\begin{thebibliography}{27}
\providecommand{\natexlab}[1]{#1}
\providecommand{\url}[1]{\texttt{#1}}
\expandafter\ifx\csname urlstyle\endcsname\relax
  \providecommand{\doi}[1]{doi: #1}\else
  \providecommand{\doi}{doi: \begingroup \urlstyle{rm}\Url}\fi

\bibitem[Zhao et~al.(2026)Zhao, Sun, Shi, and Song]{ahamoments2025}
Jiachen Zhao, Yiyou Sun, Weiyan Shi, and Dawn Song.
\newblock {Can Aha Moments Be Fake? Towards Quantifying Decorative and True
  Thinking in Chain-of-Thought}, 2026.
\newblock URL \url{https://arxiv.org/abs/2510.24941}.

\bibitem[Panickssery et~al.(2023)Panickssery, Gabrieli, Schulz, Tong, Hubinger,
  and Turner]{panickssery2023caa}
Nina Panickssery, Nick Gabrieli, Julian Schulz, Meg Tong, Evan Hubinger, and
  Alexander~Matt Turner.
\newblock {Steering Llama 2 via Contrastive Activation Addition}, 2023.
\newblock URL \url{https://arxiv.org/abs/2312.06681}.

\bibitem[Venhoff et~al.(2025)Venhoff, Arcuschin, Torr, Conmy, and
  Nanda]{venhoff2025steering}
Constantin Venhoff, Iván Arcuschin, Philip Torr, Arthur Conmy, and Neel Nanda.
\newblock {Understanding Reasoning in Thinking Language Models via Steering
  Vectors}, 2025.
\newblock URL \url{https://arxiv.org/abs/2506.18167v3}.

\bibitem[Fraser-Taliente et~al.(2026)Fraser-Taliente, Kantamneni, Ong, Mossing,
  Lu, Bogdan, Ameisen, Chen, Kishylau, Pearce, Tarng, Wu, Wu, Zhang, Ziegler,
  Hubinger, Batson, Lindsey, Zimmerman, and Marks]{frasertaliente2026nla}
Kit Fraser-Taliente, Subhash Kantamneni, Euan Ong, Dan Mossing, Christina Lu,
  Paul~C. Bogdan, Emmanuel Ameisen, James Chen, Dzmitry Kishylau, Adam Pearce,
  Julius Tarng, Alex Wu, Jeff Wu, Yang Zhang, Daniel~M. Ziegler, Evan Hubinger,
  Joshua Batson, Jack Lindsey, Samuel Zimmerman, and Samuel Marks.
\newblock Natural language autoencoders produce unsupervised explanations of
  {LLM} activations.
\newblock \emph{Transformer Circuits Thread}, 2026.
\newblock URL \url{https://transformer-circuits.pub/2026/nla/index.html}.

\bibitem[Ghandeharioun et~al.(2024)Ghandeharioun, Caciularu, Pearce, Dixon, and
  Geva]{ghandeharioun2024patchscopes}
Asma Ghandeharioun, Avi Caciularu, Adam Pearce, Lucas Dixon, and Mor Geva.
\newblock {Patchscopes: A Unifying Framework for Inspecting Hidden
  Representations of Language Models}, 2024.
\newblock URL \url{https://arxiv.org/abs/2401.06102}.

\bibitem[Turner et~al.(2023)Turner, Thiergart, Leech, Udell, Vazquez, Mini, and
  MacDiarmid]{turner2023actadd}
Alexander~Matt Turner, Lisa Thiergart, Gavin Leech, David Udell, Juan~J.
  Vazquez, Ulisse Mini, and Monte MacDiarmid.
\newblock {Steering Language Models With Activation Engineering}, 2023.
\newblock URL \url{https://arxiv.org/abs/2308.10248}.

\bibitem[Zou et~al.(2023)Zou, Phan, Chen, Campbell, Guo, Ren, Pan, Yin,
  Mazeika, Dombrowski, Goel, Li, Byun, Wang, Mallen, Basart, Koyejo, Song,
  Fredrikson, Kolter, and Hendrycks]{zou2023repe}
Andy Zou, Long Phan, Sarah Chen, James Campbell, Phillip Guo, Richard Ren,
  Alexander Pan, Xuwang Yin, Mantas Mazeika, Ann-Kathrin Dombrowski, Shashwat
  Goel, Nathaniel Li, Michael~J. Byun, Zifan Wang, Alex Mallen, Steven Basart,
  Sanmi Koyejo, Dawn Song, Matt Fredrikson, J.~Zico Kolter, and Dan Hendrycks.
\newblock {Representation Engineering: A Top-Down Approach to AI Transparency},
  2023.
\newblock URL \url{https://arxiv.org/abs/2310.01405}.

\bibitem[Chen et~al.(2024)Chen, Xu, Liang, He, Pang, Yu, Song, Liu, Zhou,
  Zhang, Wang, Tu, Mi, and Yu]{chen2024overthinking}
Xingyu Chen, Jiahao Xu, Tian Liang, Zhiwei He, Jianhui Pang, Dian Yu, Linfeng
  Song, Qiuzhi Liu, Mengfei Zhou, Zhuosheng Zhang, Rui Wang, Zhaopeng Tu,
  Haitao Mi, and Dong Yu.
\newblock {Do NOT Think That Much for 2+3=? On the Overthinking of o1-Like
  LLMs}, 2024.
\newblock URL \url{https://arxiv.org/abs/2412.21187}.

\bibitem[Muennighoff et~al.(2025)Muennighoff, Yang, Shi, Li, Fei-Fei,
  Hajishirzi, Zettlemoyer, Liang, Candès, and Hashimoto]{muennighoff2025s1}
Niklas Muennighoff, Zitong Yang, Weijia Shi, Xiang~Lisa Li, Li~Fei-Fei,
  Hannaneh Hajishirzi, Luke Zettlemoyer, Percy Liang, Emmanuel Candès, and
  Tatsunori Hashimoto.
\newblock {s1: Simple test-time scaling}, 2025.
\newblock URL \url{https://arxiv.org/abs/2501.19393}.

\bibitem[Azizi et~al.(2026)Azizi, Potraghloo, Kundu, and Pedram]{azizi2026asc}
Seyedarmin Azizi, Erfan~Baghaei Potraghloo, Souvik Kundu, and Massoud Pedram.
\newblock Activation steering for chain-of-thought compression.
\newblock In Maria Liakata, Viviane~P. Moreira, Jiajun Zhang, and David
  Jurgens, editors, \emph{Findings of the {A}ssociation for {C}omputational
  {L}inguistics: {ACL} 2026}, pages 36676--36687, San Diego, California, United
  States, July 2026. Association for Computational Linguistics.
\newblock ISBN 979-8-89176-395-1.
\newblock \doi{10.18653/v1/2026.findings-acl.1828}.
\newblock URL \url{https://aclanthology.org/2026.findings-acl.1828/}.

\bibitem[Huang et~al.(2025)Huang, Chen, Ruan, Zhang, Wei, and
  Dong]{manifoldsteering2025}
Yao Huang, Huanran Chen, Shouwei Ruan, Yichi Zhang, Xingxing Wei, and Yinpeng
  Dong.
\newblock {Mitigating Overthinking in Large Reasoning Models via Manifold
  Steering}, 2025.
\newblock URL \url{https://arxiv.org/abs/2505.22411}.

\bibitem[Chen et~al.(2025{\natexlab{a}})Chen, Zhang, Hong, Kundu, and
  Wang]{chen2025seal}
Runjin Chen, Zhenyu Zhang, Junyuan Hong, Souvik Kundu, and Zhangyang Wang.
\newblock {SEAL: Steerable Reasoning Calibration of Large Language Models for
  Free}, 2025{\natexlab{a}}.
\newblock URL \url{https://arxiv.org/abs/2504.07986}.

\bibitem[Wang et~al.(2025)Wang, Feng, Chen, Chu, Krishna, and
  Zhou]{wang2025nowait}
Chenlong Wang, Yuanning Feng, Dongping Chen, Zhaoyang Chu, Ranjay Krishna, and
  Tianyi Zhou.
\newblock {Wait, We Don't Need to ``Wait''! Removing Thinking Tokens Improves
  Reasoning Efficiency}, 2025.
\newblock URL \url{https://arxiv.org/abs/2506.08343}.

\bibitem[Yang et~al.(2025{\natexlab{a}})Yang, Si, Duan, Zhu, Zhu, Li, Chen,
  Lin, and Wang]{yang2025deer}
Chenxu Yang, Qingyi Si, Yongjie Duan, Zheliang Zhu, Chenyu Zhu, Qiaowei Li,
  Minghui Chen, Zheng Lin, and Weiping Wang.
\newblock {Dynamic Early Exit in Reasoning Models}, 2025{\natexlab{a}}.
\newblock URL \url{https://arxiv.org/abs/2504.15895}.

\bibitem[Caldarella et~al.(2026)Caldarella, Talon, Aljundi, Ricci, and
  Mancini]{caldarella2026thinkingpast}
Simone Caldarella, Davide Talon, Rahaf Aljundi, Elisa Ricci, and Massimiliano
  Mancini.
\newblock {Thinking Past the Answer: Evaluating Harmful Overthinking in Large
  Reasoning Models}, 2026.
\newblock URL \url{https://arxiv.org/abs/2606.02835}.

\bibitem[Chen et~al.(2025{\natexlab{b}})Chen, Merullo, Stolfo, and
  Pavlick]{chen2025stitching}
Alan Chen, Jack Merullo, Alessandro Stolfo, and Ellie Pavlick.
\newblock {Transferring Linear Features Across Language Models With Model
  Stitching}, 2025{\natexlab{b}}.
\newblock URL \url{https://arxiv.org/abs/2506.06609}.

\bibitem[Oozeer et~al.(2025)Oozeer, Nathawani, Prakash, Lan, Harrasse, and
  Abdullah]{oozeer2025interventiontransfer}
Narmeen Oozeer, Dhruv Nathawani, Nirmalendu Prakash, Michael Lan, Abir
  Harrasse, and Amirali Abdullah.
\newblock {Activation Space Interventions Can Be Transferred Between Large
  Language Models}, 2025.
\newblock URL \url{https://arxiv.org/abs/2503.04429}.

\bibitem[Bello et~al.(2025)Bello, Das, Zeng, Yin, and Leqi]{bello2025lrt}
Femi Bello, Anubrata Das, Fanzhi Zeng, Fangcong Yin, and Liu Leqi.
\newblock {Linear Representation Transferability Hypothesis: Leveraging Small
  Models to Steer Large Models}, 2025.
\newblock URL \url{https://arxiv.org/abs/2506.00653}.

\bibitem[{fzaffino}(2026)]{fzaffino2026foreignnla}
{fzaffino}.
\newblock Explaining {SAE} features with foreign natural language autoencoders.
\newblock LessWrong, 2026.
\newblock URL
  \url{https://www.lesswrong.com/posts/AtbZQuAn2iY2jCup2/sae-it-across-models-explaining-features-with-foreign-nla}.

\bibitem[Rafailov et~al.(2023)Rafailov, Sharma, Mitchell, Manning, Ermon, and
  Finn]{rafailov2023dpo}
Rafael Rafailov, Archit Sharma, Eric Mitchell, Christopher~D Manning, Stefano
  Ermon, and Chelsea Finn.
\newblock {Direct Preference Optimization: Your Language Model is Secretly a
  Reward Model}.
\newblock In A.~Oh, T.~Naumann, A.~Globerson, K.~Saenko, M.~Hardt, and
  S.~Levine, editors, \emph{Advances in Neural Information Processing Systems},
  volume~36, pages 53728--53741. Curran Associates, Inc., 2023.
\newblock \doi{10.52202/075280-2338}.
\newblock URL
  \url{https://proceedings.neurips.cc/paper_files/paper/2023/file/a85b405ed65c6477a4fe8302b5e06ce7-Paper-Conference.pdf}.

\bibitem[Shao et~al.(2024)Shao, Wang, Zhu, Xu, Song, Bi, Zhang, Zhang, Li, Wu,
  and Guo]{shao2024deepseekmath}
Zhihong Shao, Peiyi Wang, Qihao Zhu, Runxin Xu, Junxiao Song, Xiao Bi, Haowei
  Zhang, Mingchuan Zhang, Y.~K. Li, Y.~Wu, and Daya Guo.
\newblock {DeepSeekMath: Pushing the Limits of Mathematical Reasoning in Open
  Language Models}, 2024.
\newblock URL \url{https://arxiv.org/abs/2402.03300}.

\bibitem[Yang et~al.(2025{\natexlab{b}})Yang, Li, Yang, Zhang, Hui, Zheng, Yu,
  Gao, Huang, Lv, Zheng, Liu, Zhou, Huang, Hu, Ge, Wei, Lin, Tang, Yang, Tu,
  Zhang, Yang, Yang, Zhou, Zhou, Lin, Dang, Bao, Yang, Yu, Deng, Li, Xue, Li,
  Zhang, Wang, Zhu, Men, Gao, Liu, Luo, Li, Tang, Yin, Ren, Wang, Zhang, Ren,
  Fan, Su, Zhang, Zhang, Wan, Liu, Wang, Cui, Zhang, Zhou, and
  Qiu]{qwen2025qwen3}
An~Yang, Anfeng Li, Baosong Yang, Beichen Zhang, Binyuan Hui, Bo~Zheng, Bowen
  Yu, Chang Gao, Chengen Huang, Chenxu Lv, Chujie Zheng, Dayiheng Liu, Fan
  Zhou, Fei Huang, Feng Hu, Hao Ge, Haoran Wei, Huan Lin, Jialong Tang, Jian
  Yang, Jianhong Tu, Jianwei Zhang, Jianxin Yang, Jiaxi Yang, Jing Zhou,
  Jingren Zhou, Junyang Lin, Kai Dang, Keqin Bao, Kexin Yang, Le~Yu, Lianghao
  Deng, Mei Li, Mingfeng Xue, Mingze Li, Pei Zhang, Peng Wang, Qin Zhu, Rui
  Men, Ruize Gao, Shixuan Liu, Shuang Luo, Tianhao Li, Tianyi Tang, Wenbiao
  Yin, Xingzhang Ren, Xinyu Wang, Xinyu Zhang, Xuancheng Ren, Yang Fan, Yang
  Su, Yichang Zhang, Yinger Zhang, Yu~Wan, Yuqiong Liu, Zekun Wang, Zeyu Cui,
  Zhenru Zhang, Zhipeng Zhou, and Zihan Qiu.
\newblock {Qwen3 Technical Report}, 2025{\natexlab{b}}.
\newblock URL \url{https://arxiv.org/abs/2505.09388}.

\bibitem[Cobbe et~al.(2021)Cobbe, Kosaraju, Bavarian, Chen, Jun, Kaiser,
  Plappert, Tworek, Hilton, Nakano, Hesse, and Schulman]{cobbe2021gsm8k}
Karl Cobbe, Vineet Kosaraju, Mohammad Bavarian, Mark Chen, Heewoo Jun, Lukasz
  Kaiser, Matthias Plappert, Jerry Tworek, Jacob Hilton, Reiichiro Nakano,
  Christopher Hesse, and John Schulman.
\newblock {Training Verifiers to Solve Math Word Problems}, 2021.
\newblock URL \url{https://arxiv.org/abs/2110.14168}.

\bibitem[Hendrycks et~al.(2021)Hendrycks, Burns, Kadavath, Arora, Basart, Tang,
  Song, and Steinhardt]{hendrycks2021math}
Dan Hendrycks, Collin Burns, Saurav Kadavath, Akul Arora, Steven Basart, Eric
  Tang, Dawn Song, and Jacob Steinhardt.
\newblock {Measuring Mathematical Problem Solving With the MATH Dataset}, 2021.
\newblock URL \url{https://arxiv.org/abs/2103.03874}.

\bibitem[Kang et~al.(2026)Kang, Liu, Ma, Huang, Tan, and
  Jiang]{kang2026promptactivationdualityimprovingactivation}
Diancheng Kang, Zheyuan Liu, Ningshan Ma, Yue Huang, Zhaoxuan Tan, and Meng
  Jiang.
\newblock Prompt-activation duality: Improving activation steering via
  attention-level interventions, 2026.
\newblock URL \url{https://arxiv.org/abs/2605.10664}.

\bibitem[Hu et~al.(2021)Hu, Shen, Wallis, Allen-Zhu, Li, Wang, Wang, and
  Chen]{hu2021lora}
Edward~J. Hu, Yelong Shen, Phillip Wallis, Zeyuan Allen-Zhu, Yuanzhi Li, Shean
  Wang, Lu~Wang, and Weizhu Chen.
\newblock {LoRA: Low-Rank Adaptation of Large Language Models}, 2021.
\newblock URL \url{https://arxiv.org/abs/2106.09685}.

\bibitem[Kwon et~al.(2023)Kwon, Li, Zhuang, Sheng, Zheng, Yu, Gonzalez, Zhang,
  and Stoica]{kwon2023vllm}
Woosuk Kwon, Zhuohan Li, Siyuan Zhuang, Ying Sheng, Lianmin Zheng, Cody~Hao Yu,
  Joseph~E. Gonzalez, Hao Zhang, and Ion Stoica.
\newblock {Efficient Memory Management for Large Language Model Serving with
  PagedAttention}, 2023.
\newblock URL \url{https://arxiv.org/abs/2309.06180}.

\end{thebibliography}
\appendix
\section{How the Construction of \texorpdfstring{$N$}{N} Affects Its Behavior}
\label{construction:appendix}
One donor trace is enough to build a working direction. Directions built from
more traces, other sites, or another averaging rule point almost the same way
and shorten outputs by a similar amount. This appendix gives the evidence.

\paragraph{Variants and test set.}
All variants come from pool42. These are the 42 random correct 1.7B traces
from which we chose the donor trace (Section~\ref{construction:evidence}): 14
from GSM8K, 14 from AMC 2023, and 14 from AIME 2025, each with at least six
donor sites. Together they give 656 difference vectors $\delta_i$. We vary four
things: the number of difference vectors, the number of traces, the depth of the
sites in the trace, and the averaging rule. We test each variant on cons120: 120
questions (117 GSM8K, 3 AMC 2023) that share no question with pool42.

This is an earlier test than the main policy. It steers at every paragraph
boundary inside thinking, with $\alpha=1$, a 6,144-token budget, and
one sample per question. Mean tokens count all generated tokens, including the
final answer and runs that reach the budget. Runs that reach the budget count
as wrong.

\begin{table}[htbp]\centering\small
\caption{Construction variants on cons120 (one sample per question).
``Normalized'' averages unit difference vectors and then scales the mean to
unit length, as for the deployed $N$ (Eq.~\ref{eq:edit}). ``Raw'' averages the
unscaled difference vectors. The 297-site row uses dpairs297, the 297 traces
from which we also build $D$. Detected doubt paragraphs are the mean count per
answer, by the English marker rule of Section~\ref{construction:evidence}.
With one sample on 120 questions, accuracy differences of this size (for
example, 96.7\% vs 92.5\%) are within noise.}
\label{construction:ablation}
\begin{tabular}{lrrr}\toprule
Direction construction & Mean tokens & Accuracy & \begin{tabular}[b]{@{}r@{}}Detected doubt\\paragraphs\end{tabular}\\\midrule
No steering (base) & 2,158.3 & 92.5\% & 13.46\\
35 sites, one trace, normalized ($N$) & 1,282.7 & 96.7\% & 0.69\\
40 traces, one random site each, normalized & 1,250.8 & 95.8\% & 0.44\\
40 traces, first site of each, normalized & 1,288.8 & 92.5\% & 0.88\\
465 sites, 30 traces, normalized & 1,344.3 & 92.5\% & 1.13\\
465 sites, 30 traces, raw & 1,289.6 & 92.5\% & 0.81\\
297 sites, 297 traces & 1,239.8 & 95.8\% & 0.52\\
Random direction, same norm & 2,151.4 & 89.2\% & 10.59\\\bottomrule
\end{tabular}
\end{table}

\paragraph{Results.}
Every NLA variant cuts mean length from about 2,160 to about 1,240--1,340 tokens
and removes most detected doubt paragraphs (Table~\ref{construction:ablation}).
A random direction of the same norm barely changes either. No multi-trace variant clearly
does better than the one-trace $N$. The test is too small to rank the variants,
and we built the one-trace direction from only one trace, so other single
traces may work less well.

The variants are also close in direction. Their pairwise cosines range from
0.93 to 0.999. The deployed $N$ averages normalized differences. Averaging the
raw differences of the same 35 sites gives a direction with cosine 0.99914 to
$N$.

\paragraph{Source data and template wording.}
The donor data do not need to come from the test dataset. A direction from eight
boundaries of one GSM8K trace lowers local doubt on 14 AIME probes from 88.4\%
to 6.2\%. Here local doubt means that a short continuation starts with doubt, as
in Section~\ref{inference:local}. This test measures local doubt only, not final
accuracy.

The wording of the template matters more than these construction choices. Five
paraphrases of the continue template give directions with a mean pairwise
cosine of only 0.754. We therefore keep the template wording fixed.

\section{Question Sets and Evaluation Details}
\label{appendix:cohorts}
Table~\ref{setup:cohorts} lists every question set (\emph{cohort}) in the paper
and the short name we use for it.

\begin{table}[htbp]
\centering\footnotesize
\setlength{\tabcolsep}{3.5pt}
\caption{Cohorts used in this paper. ``Samples'' is the number of generated
samples per question and method. ``Fixed'' marks saved traces that we only read,
with no new generation. ``4+4'' means four base and four steered teacher traces
per question. The \emph{census} is an earlier screening run that sampled every
GSM8K question four times; a \emph{band} is a difficulty band from this census
(see below). Token budgets are 16,384 for gsm997, gsm1319, dev200, and dev400; 32,768
for MATH-500, confirm400, gate400, and hard128; 12,288 for cont387 and dom150;
and 6,144 for cons120. The test cohorts gsm997, MATH-500, confirm400, gate400,
sel400, and fresh1000 share no question text with each other or with dev400.
gsm1319 contains gsm997 and the 86 GSM8K test questions of dev400.
gate400 is one question set used twice: as the probe-gate test for the 1.7B and
as the hard test for the 8B students.}
\label{setup:cohorts}
\begin{tabularx}{\linewidth}{@{}l>{\raggedright\arraybackslash\hsize=1.2\hsize}Xrl>{\raggedright\arraybackslash\hsize=0.8\hsize}X>{\raggedright\arraybackslash}p{2.2cm}@{}}
\toprule
Name & Source & Size & Samples & Purpose & Used in\\
\midrule
\multicolumn{6}{@{}l}{\emph{Test cohorts}}\\
gsm997 & GSM8K test questions not used in earlier experiments; all solved 4/4 in the census & 997 & 4 & Main frozen-policy test & \S\ref{inference:frozen}\\
gsm1319 & All GSM8K test questions: gsm997 and the 322 others & 1,319 & 4 & Frozen-policy test on natural GSM8K & \S\ref{inference:frozen}\\
MATH-500 & MATH-500 (levels 1--5: 43/90/105/128/134) & 500 & 4 & Frozen-policy test; levels & \S\ref{inference:frozen}, \S\ref{limits:hard}\\
confirm400 & MATH test minus MATH-500; same level mix & 400 & 4 & Transfer test; comparators & \S\ref{inference:comparators}, \S\ref{sec:transfer}, App.~\ref{appendix:comparators}, \ref{distill:setup}\\
hard128 & AIME 2024 (30), AIME 2025 (28; the donor-trace problem and one development problem left out), AMC 2023 (40), HMMT Feb.\ 2025 (30) & 128 & 8 & Competition stress test & \S\ref{limits:hard}\\
gate400 & MATH test, levels 4 and 5 (200 each) & 400 & 4 & Probe gate (1.7B); student test (8B) & \S\ref{limits:selectivity}, \S\ref{sec:distillation}, App.~\ref{distill:setup}\\
sel400 & MATH test, levels 4 and 5 (200 each) & 400 & 4 & Redundant-only direction test & \S\ref{limits:selectivity}\\
fresh1000 & MATH test, 200 per level & 1,000 & 4 & Student test & \S\ref{sec:distillation}, App.~\ref{distill:setup}\\
pipe956 & MATH test; 1,000 planned, level 1 short by 44 & 956 & 4 & 1.7B Gated-SFT test & App.~\ref{distill:small}\\
\midrule
\multicolumn{6}{@{}l}{\emph{Development and earlier cohorts}}\\
dev200 & Part of dev400: GSM8K (157 train, 43 test); 50 per band & 200 & 1 & Policy screen & \S\ref{sec:setup}, App.~\ref{inference:delay}\\
dev400 & GSM8K (313 train, 86 test), 1 AIME 2025; 100 per band & 400 & 2--4 & Policy choice; $\alpha$ for larger models; SEAL tuning; $P$ and $-N$ & \S\ref{sec:setup}, \S\ref{sec:transfer}, \S\ref{limits:reverse}, App.~\ref{inference:delay}, \ref{appendix:comparators}\\
cont387 & dev400 minus 13 questions with no candidate answer & 387 & 2 & Continuation value & \S\ref{inference:comparators}, App.~\ref{appendix:continuation}\\
dom150 & GSM8K (116 train, 33 test), 1 AIME 2025; 50 per band, three harder bands & 150 & 1 & Earlier $N$ versus $D$ test & \S\ref{inference:frozen}\\
cons120 & 117 GSM8K, 3 AMC 2023; no pool42 question & 120 & 1 & Construction variants test & App.~\ref{construction:appendix}\\
\midrule
\multicolumn{6}{@{}l}{\emph{Fixed source-model traces}}\\
local200 & 200 correct 1.7B traces (147 GSM8K, 38 AMC 2023, 15 AIME 2025) & 400 sites & 4 or 8 & Local edit test & \S\ref{inference:local}\\
dpairs297 & 297 correct 1.7B traces (244 GSM8K, 38 AMC 2023, 15 AIME 2025) & 297 pairs & fixed & Build $D$ & \S\ref{sec:setup}, App.~\ref{construction:appendix}\\
pool42 & 42 random correct 1.7B traces (14 each: GSM8K, AMC 2023, AIME 2025), each with at least 6 donor sites & 42 & fixed & Choose the donor trace; other ways to build $N$ & \S\ref{construction:evidence}, App.~\ref{construction:appendix}\\
probe1000 & 1.7B base traces, MATH train levels 3--5, 935 questions & 1,000 & fixed & Label episodes; train probes & \S\ref{limits:selectivity}, App.~\ref{appendix:selectivity}\\
audit200 & MATH-500, 40 per level; base and VERS both correct & 200 pairs & fixed & Trace-quality audit & \S\ref{limits:hard}\\
mathtrain6588 & MATH train; no text shared with MATH-500, confirm400, fresh1000, gate400 & 6,588 & 4+4 & Student training (base and steered teacher) & \S\ref{sec:distillation}, App.~\ref{distill:setup}\\
\bottomrule
\end{tabularx}
\end{table}

\paragraph{Why the GSM8K test set is reliably solved.}
gsm997 holds the 997 GSM8K test questions not used in earlier experiments. Each
of them was solved 4 of 4 times in an earlier screening census, so gsm997 is an
easy subset, not natural GSM8K. Here is why this happened.
\begin{enumerate}[leftmargin=*,nosep]
\item The census sampled every GSM8K question four times.
\item Earlier experiments chose questions by difficulty. In total, they used
322 of the 1,319 GSM8K test questions.
\item We wanted a held-out set that is still the official GSM8K test split, so
we took all 997 remaining test questions and added no training questions.
\item The 322 used questions include all 212 test questions that the model
failed at least once in the census. So every one of the 997 remaining questions
was solved 4 of 4 times. We call them \emph{reliably solved}.
\end{enumerate}
Base accuracy on gsm997 is 98.92\%, so gsm997 describes questions that this
model reliably solves. To test natural GSM8K, we also ran all 1,319 test
questions (gsm1319, Section~\ref{inference:frozen}). Base accuracy there is
90.3\%, and 63.3\% on the 322 questions outside gsm997.

\paragraph{Development questions.}
We chose policies on dev200 and dev400 (dev200 is part of dev400). They are GSM8K
questions from both the training and the test split, plus one AIME 2025
question. They are balanced across four difficulty bands from the census:
solved in 4, in 2 or 3, in 1, or in 0 of 4 samples. So they are much harder than
gsm997: base accuracy is about 55\%. The 86 GSM8K test questions in dev400 are
among the 322 test questions used in earlier experiments. The larger models in
Section~\ref{sec:transfer} also choose their strength $\alpha$ on dev400.

\section{Policy Selection and Delay Ablation}
\label{inference:delay}
This appendix shows how we chose the strength $\alpha$ and the delay $d$ of each
direction on Qwen3-1.7B. A rule fixed in advance took the setting with the
fewest tokens whose accuracy estimate stayed within the $-2$ pp tolerance. It
chose $\alpha=1, d=256$ for $N$ and $\alpha=0.5, d=512$ for $D$. We then froze
both settings for all test runs.

The two directions are $N$ (Section~\ref{method:edit}) and $D$, the
difference-of-means direction built from dpairs297
(Section~\ref{sec:setup}). Both use the same simple policy: add the direction at
paragraph boundaries inside thinking after a fixed delay. The policy needs no
detector and does not look at candidate answers. Selection used development
questions only, in two stages (Table~\ref{setup:cohorts} describes both
cohorts):
\begin{enumerate}[leftmargin=*,nosep]
\item \emph{Screen} on dev200, with one sample per question. We ran 32
settings: four delays (0, 256, 512, 1,024) times four strengths (0.1, 0.25, 0.5,
1.0) for each of $N$ and $D$. A rule fixed in advance kept four settings per
direction.
\item \emph{Choice} on dev400, with two samples per question. We ran the eight
kept settings and two concision prompts, and the rule above chose one setting
per direction.
\end{enumerate}

\paragraph{Stage 1: the delay matters little.}
Strength drives the savings; the delay barely changes them.
Table~\ref{inference:delaytable} shows the four delays for $N$ at $\alpha=1$.
All rows use the same questions, base, and seed. $N$ saves 44--52\% at every
delay, and starting later does not give a steady accuracy gain. The estimate at
delay 256 is positive, but its interval includes zero.

\begin{table}[htbp]
\centering\small
\caption{Delay ablation for $N$ at $\alpha=1$, on dev200 (one sample per
question; base accuracy 55.0\%). Delay 0 means the first eligible paragraph
boundary, not the first generated token. Length counts all generated tokens.
Accuracy changes have paired 95\% intervals from a normal approximation;
positive values favor the intervention.}
\label{inference:delaytable}
\begin{tabular}{lrrr}
\toprule
Policy & Mean length & Tokens saved & $\Delta$ accuracy [CI], pp\\
\midrule
Base & 3,856 & \multicolumn{2}{c}{reference}\\
$N$, delay 0 & 1,981 & 48.6\% & $-4.0\ [-9.9,{+}1.9]$\\
$N$, delay 256 & 1,870 & 51.5\% & $+2.5\ [-2.6,{+}7.6]$\\
$N$, delay 512 & 2,049 & 46.9\% & $-1.0\ [-5.8,{+}3.8]$\\
$N$, delay 1,024 & 2,152 & 44.2\% & $+1.5\ [-2.3,{+}5.3]$\\
\bottomrule
\end{tabular}
\end{table}

\paragraph{Stage 2: the choice.}
Table~\ref{inference:stage2table} gives every setting of stage 2. The rule took
the setting with the fewest tokens whose accuracy point estimate was at least
$-2$ pp. Ties would go to higher accuracy, then smaller $\alpha$, then shorter
delay. For $N$, it chose $\alpha=1, d=256$, which saved 48.8\% at an estimated
$-0.4$ pp. For $D$, it chose $\alpha=0.5, d=512$, which saved 32.9\% at an
estimated $-1.9$ pp. Two settings sit exactly at $-2.0$ pp. They count as within
the tolerance, but they use more tokens, so the rule did not choose them. The
first concision prompt is the one we use as a baseline in
Section~\ref{inference:frozen}. This stage only chose the settings. We test
non-inferiority on the test cohorts (Section~\ref{inference:frozen}).

\begin{table}[htbp]
\centering\small
\caption{Stage 2 of policy selection on dev400 (two samples per question; base
accuracy 55.8\%). Chosen settings are shaded. Length counts all generated
tokens. ``Detected doubt reduced'': share of doubt paragraphs (English marker
rule, Section~\ref{construction:evidence}) that disappear relative to base.
Accuracy changes have paired 95\% intervals.}
\label{inference:stage2table}
\setlength{\tabcolsep}{4.5pt}
\begin{tabular}{lrrrr}
\toprule
Setting & Mean length & Tokens saved & \begin{tabular}[b]{@{}r@{}}Detected doubt\\reduced\end{tabular} & $\Delta$ accuracy [CI], pp\\
\midrule
Base & 3,716 & \multicolumn{3}{c}{reference}\\
\midrule
\rowcolor{black!10}$N$, $\alpha=1$, $d=256$ & 1,902 & 48.8\% & 89\% & $-0.4\ [-3.4,{+}2.7]$\\
$N$, $\alpha=1$, $d=1{,}024$ & 2,174 & 41.5\% & 73\% & $-2.0\ [-4.3,{+}0.3]$\\
$N$, $\alpha=0.25$, $d=256$ & 2,825 & 24.0\% & 52\% & $-0.6\ [-3.3,{+}2.1]$\\
$N$, $\alpha=0.25$, $d=1{,}024$ & 2,974 & 20.0\% & 43\% & $-1.1\ [-3.4,{+}1.1]$\\
\midrule
\rowcolor{black!10}$D$, $\alpha=0.5$, $d=512$ & 2,495 & 32.9\% & 72\% & $-1.9\ [-4.5,{+}0.7]$\\
$D$, $\alpha=0.5$, $d=256$ & 2,555 & 31.2\% & 77\% & $+0.1\ [-2.6,{+}2.8]$\\
$D$, $\alpha=0.5$, $d=1{,}024$ & 2,695 & 27.5\% & 61\% & $-1.6\ [-3.8,{+}0.6]$\\
$D$, $\alpha=0.1$, $d=0$ & 3,389 & 8.8\% & 25\% & $-2.0\ [-4.8,{+}0.8]$\\
\midrule
Concision prompt (Section~\ref{inference:frozen}) & 2,590 & 30.3\% & 36\% & $-1.6\ [-4.9,{+}1.6]$\\
Second concision prompt & 3,496 & 5.9\% & 10\% & $+0.1\ [-2.9,{+}3.2]$\\
\bottomrule
\end{tabular}
\end{table}

\subsection{Stability over long answers}
\label{appendix:longtrace}
Steering adds a vector at about 36 paragraph breaks in one answer. Each edited
state stays in the attention cache, so every later token reads it. Recent
work shows that such edits can add up: when the state is edited at every token,
the output becomes incoherent after a few turns
\cite{kang2026promptactivationdualityimprovingactivation}. The edits could therefore add up over a long answer
and damage the model. Then shorter
answers could simply mean a broken model that stops early. We check whether the
effect of steering changes from the start to the end of an answer.

\paragraph{Setup.} This is an earlier test. It steered at every paragraph
break with $\alpha=1$, using the direction built from 297 traces
(Appendix~\ref{construction:appendix}). It used 120 questions (111 GSM8K,
7 AMC 2023, 2 AIME 2025) with a budget of 6,144 tokens. It has three
conditions: base (no steering), steered, and a random vector of the same length
added at the same places. We split each answer into paragraphs and numbered
them: 1st, 2nd, 3rd, and so on. Then we grouped the paragraphs by position, for
example all 6th to 10th paragraphs of all answers.

\begin{table}[htbp]
\centering\small
\caption{Share of paragraphs that start with a doubt marker
(Section~\ref{construction:evidence}), by position in the answer. Each row
groups paragraphs by their number in the answer. For example, 41.8\% of the
6th to 10th paragraphs of base answers start with doubt, but only 0.4\% of
steered ones. ``Doubt removed'' is $1-\text{steered}/\text{base}$, here 99\%.
If the effect of steering faded over a long answer, the steered values would
rise in the lower rows. They stay near 0\% down to the 50th paragraph.}
\label{appendix:longtracetable}
\begin{tabular}{lrrrr}
\toprule
& \multicolumn{3}{c}{Paragraphs that start with doubt} & \\
\cmidrule(lr){2-4}
Paragraph position & Base & Steered & Random vector & Doubt removed\\
\midrule
1st--5th & 16.0\% & 0.0\% & 16.2\% & 100\%\\
6th--10th & 41.8\% & 0.4\% & 40.2\% & 99\%\\
11th--15th & 39.0\% & 0.3\% & 41.5\% & 99\%\\
16th--20th & 40.7\% & 0.0\% & 42.9\% & 100\%\\
21st--30th & 41.6\% & 0.5\% & 43.0\% & 99\%\\
31st--50th & 46.8\% & 0.0\% & 43.1\% & 100\%\\
\bottomrule
\end{tabular}
\end{table}

\paragraph{Result.} The effect of steering does not fade. Steering removes
99--100\% of detected doubt at every position, up to the 50th paragraph. The
random vector is added at the same places with the same length, but its values
stay close to base. So adding vectors at these places does not by itself change
the model, and the effect comes from the direction. Two other checks show no
sign of damage. Repetition falls over the steered answer, and it
stays flat with the random vector. Paragraph length stays about the same in all
three conditions. The edits may not add up because we steer only at paragraph
breaks, which are about 3--5\% of all tokens.

\paragraph{Direct test of the cache.} We also measured how much the edited
states in the cache change the tokens that come after them. We took 789
answers generated with the deployed $N$ ($\alpha=1$, at every paragraph break)
on hard questions that the model often fails, 1.93 million tokens in total. We ran each answer
through the model twice with the same tokens: once with the hook at the same
places as during generation, and once without it. For every token we compared
its probability in the two runs. About 1\% of tokens come right after a steered
break, and there the hook changes the probability strongly, as intended
(median ratio 793). All other tokens are affected only through the cache. For
them, the median ratio is 1.00, 99.1\% of ratios lie between 0.8 and 1.2, and
only 30 of 1.9 million exceed 10. So the edited states in the cache change the
later tokens very little.

\section{Comparison with Other Methods: Details}
\label{appendix:comparators}
This appendix gives the settings and full results of the comparison in
Section~\ref{inference:comparators}, the continuation-value test
(Appendix~\ref{appendix:continuation}), and one worked example
(Appendix~\ref{appendix:ellipse}). The comparison runs use Qwen3-1.7B on
confirm400, with four samples per question. Every method is compared with the
base model of the same run.

\paragraph{Settings.}
\begin{itemize}
\item \emph{Word ban.} The ban blocks the eight one-word doubt markers (``Wait'',
``Hmm'', ``Hold'', ``Actually'', ``Alternatively'', ``But'', ``Maybe'',
``However''). These words map to 36 token IDs. In base outputs, 96.2\% of doubt
paragraphs open with one of these tokens. The \emph{matched ban} blocks them
only as the first token of a paragraph inside thinking, from generated token
256 on. These are the same sites and schedule as VERS. The \emph{global ban}
blocks them everywhere.
\item \emph{SEAL.} We build SEAL's vector from 1,000 correct source traces, with
our sampler and answer-format prompt. We use the output of decoder block 20.
The original code edits one block earlier (block 19 in our counting). Default
SEAL adds its vector, at its own size, at every paragraph break inside
thinking, from the first generated token. We report this default strength and a
strength of 1.5 tuned on dev400. A further check puts SEAL's unit vector on
VERS's schedule at $\alpha=0.5$, so that only the vector differs from
half-strength VERS.
\item \emph{DEER.} DEER allows up to ten trial answers, uses a confidence
threshold of 0.95, and closes thinking at 80\% of the budget, as in its official
Qwen3 setting. Costs include the discarded trial answers. Our token count does
not measure the extra latency of branching.
\end{itemize}

\begin{table}[htbp]
\centering\small
\caption{Comparator results on confirm400 (four samples per question). Mean
length counts all generated tokens, including DEER's trial answers. ``Detected
doubt reduced'' is the share of doubt paragraphs, detected by the English marker
rule of Section~\ref{construction:evidence}, that disappear relative to base.
Accuracy changes and 95\% intervals are relative to the base of this run. The
SEAL vector on VERS's schedule uses $\alpha=0.5$.}
\label{inference:comparatortable}
\setlength{\tabcolsep}{4pt}
\begin{tabular}{lrrrr}
\toprule
Method & Mean length & Tokens saved & \begin{tabular}[b]{@{}r@{}}Detected doubt\\reduced\end{tabular} & $\Delta$ accuracy [CI], pp\\
\midrule
Base (accuracy 91.4\%) & 5,520 & \multicolumn{3}{c}{reference}\\
VERS, $\alpha=1$ & 3,744 & 32.2\% & 89.9\% & $-3.12\ [-4.62,{-}1.69]$\\
VERS, $\alpha=0.5$ & 3,877 & 29.8\% & 83.0\% & $-1.62\ [-3.00,{-}0.31]$\\
Matched ban & 4,151 & 24.8\% & 88.5\% & $-2.62\ [-4.00,{-}1.31]$\\
Global ban & 3,836 & 30.5\% & 99.1\% & $-5.06\ [-7.00,{-}3.25]$\\
SEAL, default & 4,617 & 16.4\% & 66.3\% & $-1.06\ [-2.38,{+}0.25]$\\
SEAL, tuned & 4,491 & 18.6\% & 82.5\% & $-1.75\ [-3.38,{-}0.25]$\\
SEAL vector, VERS schedule & 4,445 & 19.5\% & 81.2\% & $-2.25\ [-3.69,{-}0.94]$\\
DEER & 3,444 & 37.6\% & 47.3\% & $-2.94\ [-4.75,{-}1.31]$\\
\bottomrule
\end{tabular}
\end{table}

\paragraph{Pairwise comparisons.}
Half-strength VERS passes the pre-set rule against the matched ban, tuned SEAL,
and SEAL's vector on VERS's schedule (Table~\ref{inference:pairwisetable}). The
rule is the one of Section~\ref{inference:comparators}: a method passes against
another if it saves more (interval above zero) and its accuracy difference has a
lower end of at least $-2$ pp. Each row gives the savings of the two methods,
the difference in savings, and the difference in accuracy, with paired 95\%
bootstrap intervals over questions. The intervals are not adjusted for multiple
comparisons.

\begin{table}[htbp]
\centering\footnotesize
\setlength{\tabcolsep}{3.5pt}
\caption{Pairwise comparisons on confirm400, VERS minus the other method. The
difference in tokens saved is VERS's saving minus the other method's saving, in
points of base tokens. It is computed before rounding, so it can differ by 0.1
from the difference of the rounded savings. A positive accuracy difference
favors VERS.}
\label{inference:pairwisetable}
\begin{tabular}{llrrl}
\toprule
VERS & Other method & \begin{tabular}[b]{@{}r@{}}Tokens saved\\(VERS vs other)\end{tabular} & \begin{tabular}[b]{@{}r@{}}Difference in\\tokens saved [CI]\end{tabular} & $\Delta$ accuracy [CI], pp\\
\midrule
$\alpha=0.5$ & Matched ban & 29.8\% vs 24.8\% & $+4.96\ [+1.66,{+}8.40]$ & $+1.00\ [-0.38,{+}2.38]$\\
$\alpha=1$ & Matched ban & 32.2\% vs 24.8\% & $+7.38\ [+3.49,{+}11.52]$ & $-0.50\ [-2.01,{+}0.94]$\\
$\alpha=0.5$ & SEAL, tuned & 29.8\% vs 18.6\% & $+11.12\ [+7.41,{+}14.87]$ & $+0.12\ [-1.31,{+}1.62]$\\
$\alpha=0.5$ & SEAL vector, VERS schedule & 29.8\% vs 19.5\% & $+10.27\ [+6.99,{+}13.85]$ & $+0.62\ [-0.81,{+}2.06]$\\
$\alpha=0.5$ & SEAL, default & 29.8\% vs 16.4\% & $+13.39\ [+10.42,{+}16.40]$ & $-0.56\ [-2.06,{+}0.88]$\\
$\alpha=1$ & DEER & 32.2\% vs 37.6\% & $-5.43\ [-10.28,{-}0.75]$ & $-0.19\ [-1.88,{+}1.62]$\\
$\alpha=0.5$ & DEER & 29.8\% vs 37.6\% & $-7.85\ [-12.46,{-}3.24]$ & $+1.31\ [-0.31,{+}2.94]$\\
\bottomrule
\end{tabular}
\end{table}

The other rows read as follows:
\begin{enumerate}[leftmargin=*,nosep]
\item Full-strength VERS saves more than the matched ban, but it does not pass
the rule: the lower end of its accuracy interval, $-2.01$ pp, is just below
$-2$ pp.
\item DEER passes the rule against full-strength VERS.
\item Against default SEAL and DEER, half-strength VERS sits between the two.
It saves more tokens than default SEAL (29.8\% vs 16.4\%) but fewer than DEER
(37.6\%). It loses more accuracy than default SEAL (1.62 vs 1.06 pp) but less
than DEER (2.94 pp). Neither pairwise comparison passes the rule in either
direction.
\end{enumerate}

For such trade-offs, the protocol fixed one tie-break in advance: tokens per
accepted rollout in a pipeline that checks answers against the reference, tries
the method twice, and then tries base twice (Table~\ref{inference:tiebreaktable}).
By this tie-break, half-strength VERS is cheaper than default SEAL, and DEER is
cheaper than half-strength VERS.

\begin{table}[htbp]
\centering\small
\caption{Tie-break on confirm400: tokens per accepted rollout in the
check-and-retry pipeline (lower is cheaper). Coverage is 96.25\% for VERS,
SEAL, and DEER.}
\label{inference:tiebreaktable}
\begin{tabular}{lr}
\toprule
Method & Tokens per accepted rollout\\
\midrule
Base & 13,084\\
SEAL, default & 11,084\\
VERS, $\alpha=0.5$ & 9,688\\
DEER & 9,083\\
\bottomrule
\end{tabular}
\end{table}

\subsection{Continuation value}
\label{appendix:continuation}
This test asks what the reasoning after a settled answer is worth, and whether
VERS keeps that value. Earlier results showed that much reasoning happens after
an answer appears. They could not show whether this later reasoning is useless
or repairs wrong answers. Here all arms start from exactly the same text, so any
difference comes from what happens next.

We use cont387: the dev400 questions where the model gives a candidate answer.
The model thinks until its answer settles. At each paragraph boundary, we ask
for its current answer by appending \texttt{</think>} and
\texttt{\textbackslash boxed\{}; the reference answer is never used. The answer
has settled when four answers in a row agree. This happens after about 540
thinking tokens (median). From exactly this text, the model continues in five
ways:
\begin{enumerate}[leftmargin=*,nosep]
\item return the settled answer;
\item close thinking and write an answer;
\item keep thinking normally (base);
\item keep thinking with VERS;
\item keep thinking with $D$.
\end{enumerate}
Each arm has two samples and a budget of
12,288 tokens. $N$ and $D$ are added at $\alpha=1$ at every paragraph boundary
after the settle point, with no delay.

\begin{table}[htbp]
\centering\small
\caption{Continuation value on cont387. All arms start from the same text, at
the point where the answer settles. Thinking tokens are counted after this
point. Accuracy changes are relative to keeping thinking normally, with 95\%
intervals clustered by question.}
\label{inference:continuationtable}
\begin{tabular}{lrrr}
\toprule
What the model does after its answer settles & Accuracy & Thinking tokens & $\Delta$ accuracy [CI], pp\\
\midrule
(1) Return the settled answer & 39.5\% & 0 & $-13.7\ [-17.8,{-}9.6]$\\
(2) Close thinking and write an answer & 49.4\% & 0 & $-3.9\ [-7.3,{-}0.5]$\\
(3) Keep thinking normally (base) & 53.2\% & 2,388 & reference\\
\rowcolor{black!10}(4) Keep thinking with VERS ($N$) & 53.0\% & 588 & $-0.3\ [-3.2,{+}2.7]$\\
(5) Keep thinking with $D$ & 54.5\% & 1,468 & $+1.3\ [-1.3,{+}3.8]$\\
\bottomrule
\end{tabular}
\end{table}

Stopping costs accuracy, while VERS shortens the later reasoning
(Table~\ref{inference:continuationtable}). Returning the settled answer loses
13.7 pp, and closing thinking loses 3.9 pp. Closing thinking loses most on
questions whose settled answer was wrong: $-7.1$ pp $[-11.8,{-}2.3]$. VERS cuts
the later thinking by 75\% (2,388 to 588 tokens), and $D$ cuts it by 38\%.

\subsection{Example: shortening a correct solution}
\label{appendix:ellipse}
Fewer detected doubt paragraphs do not always mean fewer checks. We show one
ellipse-area problem (MATH intermediate\_algebra:133, seed 0), where every method
gives the correct answer, $15\pi$ (Table~\ref{inference:ellipsetable}).

\begin{table}[htbp]
\centering\small
\caption{One MATH test problem solved by each method (seed 0). Tokens count all
generated tokens; for DEER they include its trial answer. Doubt paragraphs are
detected by the English marker rule of Section~\ref{construction:evidence}.}
\label{inference:ellipsetable}
\begin{tabular}{lrrl}
\toprule
Method & Tokens & \begin{tabular}[b]{@{}r@{}}Detected doubt\\paragraphs\end{tabular} & What happens\\
\midrule
Base & 3,682 & 13 & finds $15\pi$, then checks again and again\\
VERS, $\alpha=0.5$ & 2,603 & 0 & keeps one geometric check, with no marker\\
Matched ban & 3,147 & 0 & keeps checking, in Chinese\\
SEAL, default & 2,746 & 4 & keeps some checks\\
DEER & 2,153 & 1 & stops after the area calculation\\
\bottomrule
\end{tabular}
\end{table}

Two methods remove all detected doubt paragraphs, but in different ways. VERS
still makes one geometric check of the result, and this check has no marker.
The matched ban blocks the English words, but the model keeps checking in
Chinese, which the English rule does not see:
\cjktext{但是，让我再仔细检查一遍} (``But let me check carefully once more'').
DEER stops right after the area calculation, with one detected doubt paragraph
before it exits.

\section{Transfer Details}
\label{appendix:transfer}
This appendix gives the details behind Section~\ref{sec:transfer}. Steering
transfer (Section~\ref{transfer:results}) uses only the maps and $N$. The
interface tests (Section~\ref{transfer:fidelity}) also use the frozen AV and AR.

\paragraph{Pair sets and maps.}
We record both residual streams at the same tokens of the same text. The text is
web documents plus reasoning traces from both models, read at paragraph
boundaries and at random positions. The pair sets have 46,671 positions for the
8B, 46,688 for the 30B-A3B, and 46,677 for the 235B-A22B. We split them by
document or by question, so all traces of one question fall in the same split.
We fit $E$ and $U$ separately and choose the ridge strength on validation data.
For the 8B and the 30B-A3B, the maps used to read and rebuild states
(Table~\ref{transfer:tab-fidelity} and the reading test) are fit with 24,000
more web and chat rows.

\paragraph{Layers.}
The target layer keeps the source layer's relative depth, about 70\% of the
model: $l_t\approx(20/28)L_t$, where $l_t$ is a zero-indexed block and $L_t$ is
the number of blocks. This gives layer 26 of 36 (8B), 34 of 48 (30B-A3B), and
67 of 94 (235B-A22B). In the mixture-of-experts models, we read and steer the
block's residual output after the expert outputs are added together.

\paragraph{Reading and rebuilding a target state.}
Reading and rebuilding a target state uses the whole interface:
\begin{equation}
 e_t=\mathrm{AV}\!\left(1000\frac{E(h_t)}{\|E(h_t)\|_2}\right),\qquad
 \hat h_t=U\!\left(m_s\frac{\mathrm{AR}(e_t)}{\|\mathrm{AR}(e_t)\|_2}\right).
 \label{transfer:eq-interface}
\end{equation}
The AV reads the mapped state at norm 1,000, as in its training, and writes the
explanation $e_t$. $U$ was fit on source states, so we rescale the AR output to
$m_s=894.97$ before we apply $U$. This is the median source-state norm in the
evaluation data. Source replay and the no-text cycle use the same rescaling.
For FVE, we scale both $\hat h_t$ and $h_t$ to unit length, as in
Eq.~\ref{eq:fve}. Steering uses none of these steps: it only maps $N$ through
$W_U$ and normalizes again. $N$ is a difference of two AR reconstructions, so
the bias $b_U$ cancels.

\begin{table}[htbp]
\centering\small
\begin{adjustbox}{max width=\linewidth}
\begin{tabular}{llrr}
\toprule
Path & Mapping chain & 8B & 30B-A3B\\
\midrule
Adapted & $h_t\to E\to\mathrm{AV}\to e_t\to\mathrm{AR}\to U\to\hat h_t$ & 0.334 & 0.386\\
Source replay & $h_s\to\mathrm{AV}\to e_s\to\mathrm{AR}\to U\to\hat h_t$ & 0.318 & 0.374\\
Ridge from $h_s$ & $h_s\to U\to\hat h_t$ (no text) & 0.437 & 0.500\\
MLP from $h_s$ & $h_s\to$ affine map $+$ MLP (4,096 hidden units) $\to\hat h_t$ (no text) & 0.465 & 0.512\\
No-text cycle & $h_t\to E\to U\to\hat h_t$ (no AV or AR) & 0.584 & 0.686\\
Shuffled & Adapted, with explanations shuffled across positions & $-0.612$ & $-0.613$\\
\midrule
Source NLA (reference) & $h_s\to\mathrm{AV}\to e_s\to\mathrm{AR}\to\hat h_s$ & \multicolumn{2}{c}{0.632}\\
\bottomrule
\end{tabular}
\end{adjustbox}
\caption{FVE on 10,000 fresh web and chat states. $h_s$ is the source state on
the same text. Target rows are scored against $h_t$; the reference row is
scored against $h_s$. Paths through the AR rescale its output to the median
source norm before $U$. Prediction and target are scaled to unit length before
scoring. Ridge and MLP are the best maps we tried, not upper bounds.}
\label{transfer:tab-fidelity}
\end{table}

\paragraph{What the fidelity table shows.}
The text step loses the most. The no-text cycle, which starts from $h_t$, scores
higher than the ridge and MLP predictors, which start from $h_s$. The Adapted
path scores slightly above Source replay. Shuffled explanations give a negative
FVE, so the explanations carry information about their own position. On 456
reasoning positions, FVE is lower than on web and chat text: 0.233 for the
Adapted path on the 8B, 0.275 on the 30B-A3B, and 0.304 for the source NLA.

\paragraph{The reading test.}
We use about 964 held-out paragraph boundaries per target. At each one, the
target and the 1.7B read the same text. Each model has its own probability that
the next paragraph starts with doubt. The test asks: where the two models
disagree, does the difference between their explanations follow the
disagreement? We score each explanation for doubt in three ways: a doubt-word
cue in its last paragraph, the projection $\langle\mathrm{AR}(e),N\rangle$, and
a blind judge that reads only the explanation. The test statistic is the
Spearman correlation $\rho$ between the score difference (target explanation
minus 1.7B explanation) and the probability difference. By a pre-stated rule,
the test passes if the 95\% interval, clustered by question, excludes zero.

Both targets pass (Table~\ref{transfer:tab-reading}).

\begin{table}[htbp]
\centering\small
\begin{tabular}{lllll}
\toprule
& \multicolumn{2}{c}{Spearman $\rho$ [95\% CI]} & \multicolumn{2}{c}{Behavioral AUC, target / 1.7B}\\
\cmidrule(lr){2-3}\cmidrule(lr){4-5}
Doubt score & 8B & 30B-A3B & 8B & 30B-A3B\\
\midrule
Doubt-word cue & $+0.15$ [0.08, 0.23] & $+0.11$ [0.04, 0.18] & 0.736 / 0.687 & 0.685 / 0.686\\
$\langle\mathrm{AR}(e),N\rangle$ & $+0.19$ [0.12, 0.27] & $+0.12$ [0.04, 0.21] & 0.780 / 0.754 & 0.707 / 0.707\\
Blind judge & $+0.09$ [0.02, 0.16] & $+0.07$ [0.00, 0.14] & 0.755 / 0.729 & 0.710 / 0.693\\
\bottomrule
\end{tabular}
\caption{The reading test. $\rho$ compares the doubt-score
difference of the two explanations with the difference in the two models'
probability of doubting next. The behavioral AUC (see text) compares
explanations of the target state with explanations of the 1.7B state on the
same text.}
\label{transfer:tab-reading}
\end{table}

The \emph{behavioral AUC} is a second statistic. It measures how well an
explanation's doubt score predicts whether the target will likely doubt next
(target probability above 0.5). We compute it on the quarter of boundaries with
the largest disagreement in each direction. Explaining the target state instead
of the 1.7B state raises this AUC by 0.026--0.049 on the 8B and by about
0.00--0.02 on the 30B-A3B.

A no-text control runs the same correlation without the AV or the AR. It uses
the projection on $N$ of the mapped target state minus that of the 1.7B state.
It gives $\rho=+0.38$ on the 8B and $+0.17$ on the 30B-A3B, more than any
explanation score. So the mapped states keep more of the target-specific signal
than the explanations do.

\paragraph{Building the direction on the target.}
We also built the direction on the target itself. The AV reads the target's own
doubt states through $E$, we apply the same continue edit, and we map the AR
differences back with $W_U$. This gives almost the same direction as mapping
$N$: cosine 0.958 on the 8B and 0.950 on the 30B-A3B. On dev400 at $\alpha=0.75$, both directions save
49.1\% of tokens. For this edit, building the direction on the target adds
little.

\paragraph{Mapped 1.7B $D$ on the 30B-A3B.}
On the 30B-A3B, the mapped 1.7B direction $D$ does not pass the development
accuracy rule (Section~\ref{transfer:method}) at any tested strength (best $-1.9$ pp). So we did not run it on
confirm400 (Table~\ref{transfer:tab-controls} reports its development result).

\paragraph{Two 1.7B results on confirm400.}
The 1.7B row of Table~\ref{transfer:tab-behavior} and the 1.7B results in
Table~\ref{inference:comparatortable} come from two separate runs. We compare
each run only with its own base: 90.2\% in the transfer run and 91.4\% in the
comparator run.

\section{Hard-Problem and Selectivity Details}
\label{appendix:selectivity}
This appendix gives the details behind Section~\ref{sec:limits}: VERS on hard
problems at each model size, the two selective constructions, and the two
directions that steer toward doubt ($P$ and $-N$).

\subsection{Hard problems at each model size}
\begin{figure}[htbp]
\centering
\includegraphics[width=0.86\linewidth]{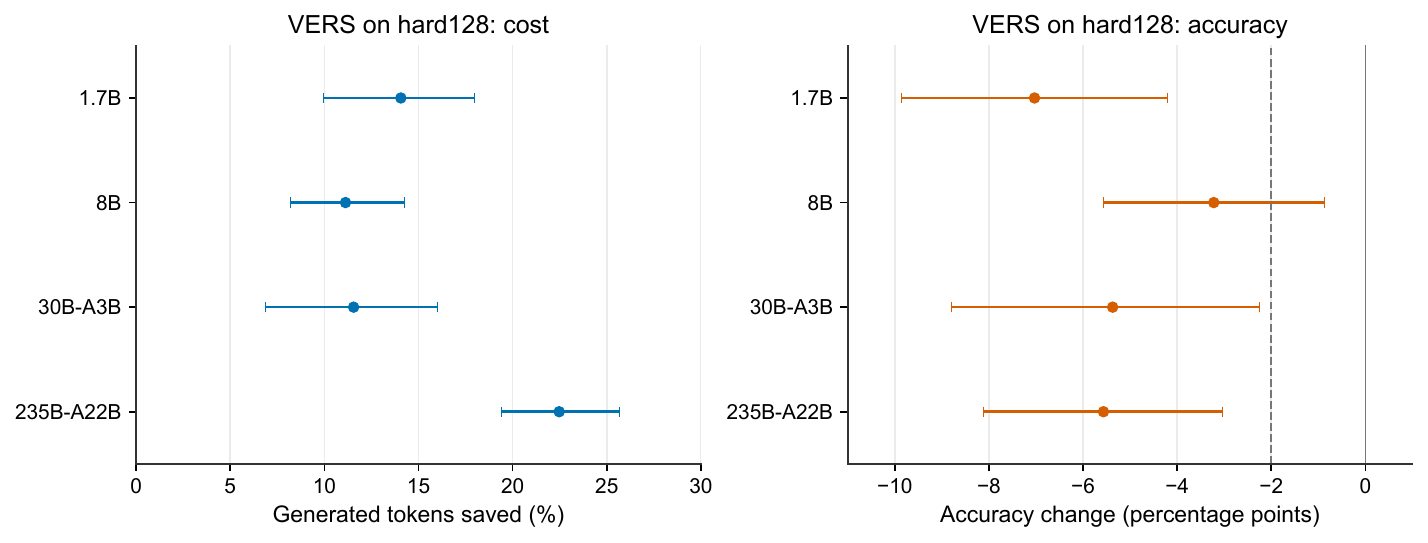}
\caption{VERS at its primary strength on hard128, for each model size. Left: generated tokens saved (\%). Right: accuracy change from the same model's base (pp). Error bars are pointwise 95\% intervals from a paired bootstrap over questions. The dashed line marks the $-2$ pp non-inferiority margin.}
\label{fig:competition-scale}
\end{figure}

Figure~\ref{fig:competition-scale} shows VERS on hard128 at each model size
(Section~\ref{limits:hard}).

\subsection{Which doubt is useful?}
To steer selectively, we first need to know which doubt helps. We label each
doubt with its effect on the answer.

A \emph{doubt episode} starts at a doubt paragraph. It ends at whichever comes
first: the next doubt paragraph, four paragraphs later, or the end of thinking.
We stop the model twice, right before the episode and at its end, and each time
force it to give its current answer. The two answers give the label
(Table~\ref{sel:labels}). For example, if the answer is wrong before the
episode and right after it, the episode is \emph{productive}: the check fixed
an error. If the answer is right before and stays the same, the episode is
\emph{redundant}: the check changed nothing.

We label all episodes in probe1000: 1,000 base 1.7B rollouts on 935 MATH
training questions of levels 3 to 5. Each doubt paragraph starts one episode,
which gives 43,539 episodes.

\begin{table}[htbp]
\centering\small
\begin{tabular}{llrr}
\toprule
Label & Forced answer, before $\rightarrow$ after & Episodes & Share\\
\midrule
Productive & wrong $\rightarrow$ right & 769 & 1.8\%\\
Redundant & right $\rightarrow$ right, same answer & 18,026 & 41.4\%\\
Failed & wrong $\rightarrow$ wrong & 24,188 & 55.6\%\\
Harmful & right $\rightarrow$ wrong & 520 & 1.2\%\\
Other & right $\rightarrow$ right, different form & 10 & 0.02\%\\
Excluded & no usable answer before & 26 & 0.06\%\\
\bottomrule
\end{tabular}
\caption{Labels of the 43,539 doubt episodes in probe1000. Useful doubt is rare:
only 1.8\% of episodes fix a wrong answer.}
\label{sel:labels}
\end{table}

These labels only approximate how useful a check is. A productive episode
improves the forced answer right away, but the model might have found the right
answer another way. A redundant episode may still check a condition or prevent
a later error. Most episodes are failed: the answer is wrong before and after.
Our tests of selectivity say nothing about failed episodes.

\subsection{Two probes}
Can the model's state tell useful doubt from unneeded doubt? We fit two
logistic-regression probes on the layer-20 state in probe1000
(Table~\ref{sel:probes}). Probe A asks, just before a doubt paragraph: will
this episode be productive or redundant? Probe B asks, at every paragraph
boundary: does a productive episode start right after this boundary? Probe A
reaches AUC 0.75 and probe B reaches AUC 0.84, so the state carries some
signal. For probe B, the chosen regularization strength was at one end of the
grid that we fixed in advance, and we did not search further.

\begin{table}[htbp]
\centering\small
\begin{tabularx}{\linewidth}{@{}l>{\raggedright\arraybackslash}X>{\raggedright\arraybackslash}X@{}}
\toprule
& Probe A & Probe B\\
\midrule
Read at & the boundary just before a doubt paragraph & every paragraph boundary inside thinking\\
Predicts & productive or redundant episode & a productive episode starts next\\
Examples & 18,795 episodes (769 productive, 18,026 redundant); failed and harmful left out & 150,501 boundaries; 769 positive (about 0.5\%)\\
Evaluation & 5-fold cross-validation, grouped by question & held-out questions (80/20 split by question)\\
AUC [95\% CI] & 0.75 [0.718, 0.782] & 0.84 [0.809, 0.873]\\
Use & selective directions & the gate\\
\bottomrule
\end{tabularx}
\caption{The two logistic-regression probes on the layer-20 state, both fit on
probe1000.}
\label{sel:probes}
\end{table}

A high AUC does not make a good gate when positives are this rare. What matters
is how the probe behaves at the threshold the gate uses. The gate fires at the
10\% of boundaries with the highest probe B scores. On held-out base traces, it
then catches 36.6\% [27.2, 47.7] of the boundaries before a productive episode.
Firing at random at the same rate would catch 10\%. But only about 0.5\% of all
boundaries come before a productive episode, and the gate fires at 10\%. So
most of its firings cannot come before a productive episode. We did not
measure precision directly.

\subsection{Selective directions}
The first construction changes the direction. We rebuilt the steering
direction from probe1000 boundaries in three ways: from boundaries before
redundant doubt only, before productive doubt only, and before any doubt. The
three directions are almost the same: the cosine between any two of them is
0.98--0.99. So leaving out useful doubt does not change the direction. We
tested the redundant-only direction on sel400 (Table~\ref{limits:seltable}). It
saves about as many tokens as $N$ and loses about as much accuracy.

\subsection{The probe gate}
The second construction keeps $N$ but decides where to apply it. During
generation, we read probe B at each boundary. When its score is above the
threshold, the gate does not steer at this boundary and at the next three. The
idea is to leave the model free when a useful check may start.

To test whether the gate skips the \emph{right} boundaries, we compare it with
random skipping. Random skipping also skips blocks of four boundaries, but it
starts a block at a random boundary, with probability 0.0804. We chose this
probability on 50 training questions, so that both arms leave about the same
share of boundaries unsteered (27.8\% for the gate, 28.2\% for random
skipping). So the two arms differ only in where the skipped blocks start. The
gate starts them where probe B fires, often at the start of a run of doubt.
During steering, probe B fired at 12.1\% of the boundaries where steering could
act, a little more often than the 10\% on base traces.

We test both arms on gate400. The gate reaches 86.75\% accuracy, and random
skipping reaches 87.75\%. The change is $-1.0$ pp [$-2.69$, $+0.62$], so the
interval includes zero. The gate saves 20.1\% of tokens and random skipping
29.5\%; the difference is $-9.34$ points of base tokens [$-12.37$, $-6.64$]. So
the gate does not keep more accuracy, and it saves fewer tokens.

Table~\ref{sel:episodes} shows why. The gate keeps twice as many productive
episodes per rollout as random skipping (0.48 vs 0.24). But it also keeps 2.6
times as many redundant ones (7.595 vs 2.91). So it keeps 0.063 productive
episodes per redundant one, which is fewer than random skipping (0.082). Each
time the gate fires, a whole run of doubt can continue, and most of this doubt
is redundant.

\begin{table}[htbp]
\centering\small
\begin{tabular}{lrr}
\toprule
Steering & Productive & Redundant\\
\midrule
Base (no steering) & 0.735 & 21.125\\
VERS ($N$) & 0.115 & 1.255\\
Random skipping & 0.24 & 2.91\\
Gate with probe B & 0.48 & 7.595\\
\bottomrule
\end{tabular}
\caption{Doubt episodes per rollout on gate400, counted on a sample of 200
rollouts per arm, with the labels of Table~\ref{sel:labels}. The gate keeps
more productive episodes than random skipping, but it keeps even more redundant
ones.}
\label{sel:episodes}
\end{table}

\subsection{The doubt-edit direction \texorpdfstring{$P$}{P}}
Section~\ref{limits:reverse} builds $P$ with the same code as $N$. To check
this code, we used it to rebuild $N$. The result $N'$ has $\cos(N',N)=0.996$,
so the code reproduces $N$.

The doubt edit gives a weaker and less consistent signal than the continue
edit. At each donor site, it changes the AR output less (mean norm 200 vs 288).
Its changes also agree less across the 35 sites: the mean cosine between each
site's change and the average change is 0.69 for the doubt edit and 0.80 for
the continue edit. In the local test of Section~\ref{inference:local}, the doubt
edit at boundaries right before ``Wait'' left the doubt share close to base
(96.7\% vs 96.5\%).

\subsection{The reversed direction \texorpdfstring{$-N$}{-N}}
$-N$ is the deployed $N$ with its sign flipped, used with the same schedule
(delay 256). We ran it on dev400 (two samples per question) together with a
new base run, and compare it with that base. Table~\ref{appendix:negntable}
gives the full results.

\begin{table}[htbp]
\centering\small
\caption{$-N$ on dev400, compared with base from the same run (base: 54.2\%
accuracy, 3,850 tokens, 28.7 detected doubt paragraphs per rollout, 0.75\% of
outputs at the token limit). Intervals are 95\% bootstrap intervals over
questions.}
\label{appendix:negntable}
\begin{tabular}{rrrrr}
\toprule
$\alpha$ & $\Delta$ accuracy, pp & Longer output & Doubt paragraphs & At token limit\\
\midrule
0.25 & $-0.62$ $[-3.50,{+}2.25]$ & 35.2\% $[29.2,41.7]$ & 53.2 & 1.25\%\\
0.5 & $-1.38$ $[-4.25,{+}1.50]$ & 73.2\% $[65.2,82.3]$ & 79.6 & 4.00\%\\
1.0 & $-5.00$ $[-8.00,{-}2.38]$ & 229.8\% $[208.5,254.2]$ & 223.2 & 55.25\%\\
\bottomrule
\end{tabular}
\end{table}

\section{Offline Collection and Student Training Details}
\label{distill:setup}

\paragraph{Offline data collection.}
The steering vector can create training data without being used at
deployment. We test this on confirm400 (400 MATH test questions, see
Appendix~\ref{appendix:cohorts}) with the steered 8B and 30B-A3B models. We use
a \emph{steered-first} collection policy. For each question, it first generates
two steered traces. Only if both are wrong does it generate two base traces; we
call this step the \emph{fallback}. It then keeps the shortest correct trace
among all traces it generated. The base comparator has the same budget, but
its first two traces are also base traces. \emph{Coverage} is the share of
questions that get at least one correct trace. \emph{Tokens per accepted
example} is the total number of generated tokens divided by the number of
questions that get a correct trace. The total includes tokens spent on
questions with no correct trace. Coverage stays at 99.75\% for both the 8B and
the 30B-A3B. Tokens per accepted example fall by 29.5\%. This selection uses
gold answers, so it works only offline.

A blind judge also checked trace quality. It read one base/steered pair per
question, chosen where both traces were correct. On both models, the share of
traces judged acceptable as teaching examples stayed within the pre-stated
5 pp margin. Some finer error measures got worse, for example missed required
conditions on 8B level-4 problems.

\paragraph{Natural and steered traces.}
The 8B writes traces for 6,588 MATH training questions (mathtrain6588).
\emph{Natural} traces are written with the steering hook off. \emph{Steered}
traces are written with the hook on, by the hooked teacher. The teacher uses
the 8B steering policy of Section~\ref{sec:transfer} (layer 26, $\alpha=0.75$,
delay 256). For each question, the 8B writes four natural and four steered
traces. A trace is \emph{usable} if it is correct, finished, and fits in the
16,384-token training context. Wrong traces, used only as rejected DPO traces,
must also fit in this context.

\paragraph{SFT data.}
After the filter, 6,268 questions have at least one usable natural trace and
one usable steered trace. Both SFT students use this same question set.
Short-SFT learns each question's shortest usable natural trace. Steered-SFT
learns its shortest usable steered trace. The Short-SFT data has 25.50 million
completion tokens, and the Steered-SFT data has 16.75 million.

\paragraph{DPO data.}
DPO-natural and DPO-steered~\cite{rafailov2023dpo} use three types of
preference pairs (Table~\ref{distill:tab-pairs}). Each question gives at most
one pair of each type. A length pair counts only if the rejected trace is at
least 1.5 times as long as the preferred one. Both arms have the same number of
pairs of each type: we randomly drop pairs from the arm that has more, with a
fixed seed. After this step, each arm has 2,477 pairs: 2,342 length pairs, 89
correctness pairs, and 46 anti-premature pairs. Correctness and
anti-premature pairs are rare because the 8B seldom writes a wrong trace that
fits in the training context.

\begin{table}[htbp]
\centering\small
\begin{tabularx}{\linewidth}{@{}l>{\raggedright\arraybackslash}X>{\raggedright\arraybackslash}X>{\raggedright\arraybackslash}X@{}}
\toprule
Pair type & Preferred trace & Rejected trace, DPO-natural & Rejected trace, DPO-steered\\
\midrule
Length & shortest correct trace of the arm (natural or steered) & longest correct natural trace & longest correct natural trace\\
Correctness & the length pair's preferred trace (if there is no length pair, a random correct trace of the arm) & longest wrong natural trace & longest wrong natural trace\\
Anti-premature & a random correct natural trace (the same in both arms) & shortest wrong natural trace that is not longer & shortest wrong steered trace that is not longer\\
\bottomrule
\end{tabularx}
\caption{The three DPO pair types. ``Correct'' means usable (correct, finished,
and within 16,384 tokens). The arms differ in two places. In length and
correctness pairs, DPO-steered prefers a steered trace. In anti-premature pairs,
DPO-steered rejects a wrong steered trace. Anti-premature pairs teach the
student not to stop too early with a wrong answer. Because of them, the steered
arm also contains some natural traces as the preferred trace.}
\label{distill:tab-pairs}
\end{table}

\paragraph{Training.}
All 8B students use LoRA~\cite{hu2021lora} with rank 64, LoRA alpha 128, and dropout
0.05 on all linear layers. They train for one epoch with one training seed.
SFT uses batches of 64 sequences, and DPO uses batches of 32 pairs, the sigmoid
loss with $\beta=0.1$, and the base model as reference. For each method, we
choose the learning rate on the natural-data arm and use the same rate for the
steered arm. Each candidate rate trains for 20 steps, and we take the rate with
the lowest held-out loss. This gives $10^{-4}$ for SFT and $5\times10^{-6}$ for DPO. We did not
fully fine-tune the 8B: on the 1.7B, LoRA and full fine-tuning gave the same
verdict on every comparison of steered with natural data
(Appendix~\ref{distill:small}).

\paragraph{Evaluation.}
Students run in plain vLLM with no hook and four samples per question. The two
test sets are fresh1000 (1,000 MATH test questions, 200 per level) and gate400
(400 level-4/5 MATH test questions). Neither set shares question text with the
training questions. The pre-set rule has two parts, and both must hold on both
sets. First, the token saving must have a 95\% interval that excludes zero.
Second, the lower end of the 95\% interval of the accuracy change must be at or
above $-2$ pp. The hooked teacher needs the hook, so we run it in our steering
code and compare it only with a base run in the same code.

\subsection{Student comparisons}
\label{distill:results}

Steered-SFT learns the teacher's shorter style
(Table~\ref{distill:tab-primary}). On fresh1000, detected doubt paragraphs per
trace fall from 21.5 for the base model to 1.7. Steered-SFT saves 33.2\% of
tokens on fresh1000 and passes the accuracy part of the rule there. It saves
28.7\% on gate400 but does not pass the accuracy part. The hooked teacher loses
much less accuracy than its student (last row of
Table~\ref{distill:tab-comparisons}).

\begin{table}[htbp]
\centering\small
\begin{tabular}{llrrl}
\toprule
Set & Student & Accuracy & Mean tokens & $\Delta$acc vs base, pp [95\% CI]\\
\midrule
fresh1000 & Base & 97.70\% & 4,578 & reference\\
& Short-SFT & 97.18\% & 4,568 & $-0.52$ [$-1.03,{-}0.07$]\\
& Steered-SFT & 96.85\% & 3,060 & $-0.85$ [$-1.35,{-}0.35$]\\
& DPO-natural & 97.58\% & 4,188 & $-0.12$ [$-0.50,{+}0.25$]\\
& DPO-steered & 97.30\% & 4,136 & $-0.40$ [$-0.80,{-}0.02$]\\
\midrule
gate400 & Base & 96.44\% & 6,686 & reference\\
& Short-SFT & 95.56\% & 6,548 & $-0.88$ [$-1.81,0.00$]\\
& Steered-SFT & 95.00\% & 4,768 & $-1.44$ [$-2.44,{-}0.44$]\\
& DPO-natural & 96.19\% & 6,174 & $-0.25$ [$-1.12,{+}0.62$]\\
& DPO-steered & 95.94\% & 6,037 & $-0.50$ [$-1.44,{+}0.44$]\\
\bottomrule
\end{tabular}
\caption{8B students without a hook. Many evaluation samples do not replace
independent training seeds.}
\label{distill:tab-primary}
\end{table}

The result depends on the natural-data recipe we compare against
(Table~\ref{distill:tab-comparisons}). Against Short-SFT, Steered-SFT saves
33.0\% of tokens on fresh1000 and 27.2\% on gate400. Its accuracy change is
$-0.32$ pp on fresh1000 and $-0.56$ pp on gate400, and both intervals pass the
accuracy part of the rule. Against DPO-natural, its accuracy change on gate400
is $-1.19$ pp [$-2.19$, $-0.25$], so it does not pass the accuracy part there.
DPO-natural already gives most of the small DPO-steered saving. DPO-steered is
1.24\% shorter than DPO-natural on fresh1000, but this interval includes zero,
so it does not pass the token part of the rule.

\begin{table}[htbp]
\centering\small
\begin{tabular}{lll}
\toprule
Paired comparison & fresh1000 [95\% CI] & gate400 [95\% CI]\\
\midrule
Steered-SFT vs Short-SFT, $\Delta$acc (pp) & $-0.32$ [$-0.85,{+}0.23$] & $-0.56$ [$-1.62,{+}0.44$]\\
Steered-SFT vs DPO-natural, $\Delta$acc (pp) & $-0.72$ [$-1.22,{-}0.22$] & $-1.19$ [$-2.19,{-}0.25$]\\
DPO-steered vs DPO-natural, $\Delta$tokens & $-1.24$\% [$-2.79,{+}0.37$] & $-2.21$\% [$-4.45,{-}0.02$]\\
Hooked teacher vs its base, $\Delta$acc (pp) & $-0.02$ [$-0.50,{+}0.45$] & $-0.31$ [$-1.31,{+}0.69$]\\
\bottomrule
\end{tabular}
\caption{Main paired comparisons. The hooked teacher runs with the hook and has
its own base run, so its row is a diagnostic comparison only.}
\label{distill:tab-comparisons}
\end{table}

The design does not separate three possible causes of the accuracy loss: the
training itself, the choice of shortest traces, and steering. Short-SFT also
loses accuracy without real shortening, which shows this mix. Steered-SFT also
shortens traces on GSM8K, which is not in its training data: it saves 44.7\% of
tokens, with an accuracy change of $-0.36$ pp. On AIME/AMC it saves 15.7\%,
with an accuracy change of $-4.46$ pp. We ran no reasoning-quality audit of the
students and no second training seed.

\subsection{Earlier 1.7B students}
\label{distill:small}
Before the 8B study, we ran the same kind of study on Qwen3-1.7B
(Table~\ref{distill:tab-small}). The teacher is the 1.7B model with VERS
($\alpha=1$, $d=256$). The training pool has 13,598 GSM8K and MATH training
questions. As for the 8B, we chose the learning rate of each method on its
natural-data version, with 20-step tests of held-out loss over five values. We
ran three rounds:
\begin{enumerate}
\item The \emph{main round} trains Short-SFT, Steered-SFT, DPO-natural, and
DPO-steered, each with LoRA and with full fine-tuning. As for the 8B,
Short-SFT and Steered-SFT train on the shortest correct trace of each question.
\item \emph{SFT$\to$DPO} first fine-tunes on easy questions, then runs DPO on
hard questions, with pairs built only from correct and wrong answers. It uses
LoRA, and it has a version trained on natural traces.
\item \emph{Gated-SFT} trains on steered traces, but uses base traces wherever
steering lost a correct answer.
\end{enumerate}

\begin{table}[htbp]\centering\small
\caption{Hook-free 1.7B students compared with the base model, on two test sets
per round. $\Delta$ accuracy is in pp with a 95\% interval. Negative token
savings mean longer outputs. As in Table~\ref{distill:tab-summary}, shading
marks the student with the largest token saving in the main round
(Steered-SFT), and $^\dagger$ marks students that pass the $-2$ pp accuracy
rule on both sets. Gated-SFT was tested on one set only. Four
samples per question; one training seed.}
\label{distill:tab-small}
\setlength{\tabcolsep}{4pt}
\begin{adjustbox}{max width=\linewidth}
\begin{tabular}{@{}llrrrr@{}}\toprule
& & \multicolumn{2}{c}{First set} & \multicolumn{2}{c}{Second set}\\
\cmidrule(lr){3-4}\cmidrule(lr){5-6}
Student & Tuning & $\Delta$ accuracy & Tokens saved & $\Delta$ accuracy & Tokens saved\\\midrule
\multicolumn{6}{@{}l}{\emph{Main round. First set: MATH-500; second set: confirm400}}\\
Short-SFT & full & $-0.4\ [-1.8,{+}0.8]$ & 0.3\% & $-1.6\ [-3.1,{-}0.1]$ & $-1.4$\%\\
Short-SFT & LoRA & $-0.7\ [-1.9,{+}0.5]$ & 0.2\% & $-1.3\ [-2.7,{+}0.0]$ & $-2.9$\%\\
\rowcolor{black!10}Steered-SFT & full & $-2.8\ [-4.2,{-}1.4]$ & 29.2\% & $-2.8\ [-4.2,{-}1.2]$ & 25.5\%\\
\rowcolor{black!10}Steered-SFT & LoRA & $-2.3\ [-3.6,{-}1.1]$ & \textbf{32.8\%} & $-3.6\ [-5.4,{-}1.9]$ & \textbf{28.9\%}\\
DPO-natural$^\dagger$ & full & $-1.0\ [-2.0,{+}0.0]$ & 25.3\% & $+0.1\ [-1.2,{+}1.4]$ & 23.0\%\\
DPO-natural & LoRA & $-1.5\ [-2.6,{-}0.2]$ & 23.8\% & $-1.4\ [-2.9,{+}0.1]$ & 23.6\%\\
DPO-steered$^\dagger$ & full & $-0.1\ [-1.2,{+}1.1]$ & 18.1\% & $+0.9\ [-0.5,{+}2.2]$ & 14.9\%\\
DPO-steered$^\dagger$ & LoRA & $+0.8\ [-0.4,{+}1.9]$ & 18.9\% & $-0.4\ [-1.6,{+}0.8]$ & 16.6\%\\
\midrule
\multicolumn{6}{@{}l}{\emph{SFT$\to$DPO. First set: fresh1000; second set: MATH-500}}\\
Natural SFT$\to$DPO$^\dagger$ & LoRA & $-0.4\ [-1.1,{+}0.4]$ & $-12.2$\% & $+0.0\ [-1.2,{+}1.2]$ & $-13.9$\%\\
Steered SFT$\to$DPO & LoRA & $-1.8\ [-2.6,{-}1.1]$ & 29.3\% & $-1.8\ [-3.0,{-}0.4]$ & 23.6\%\\
\midrule
\multicolumn{6}{@{}l}{\emph{Gated-SFT. First set: pipe956 (Appendix~\ref{appendix:cohorts})}}\\
Gated-SFT & full & $-1.5\ [-2.4,{-}0.6]$ & 32.5\% & \multicolumn{2}{c}{not tested}\\
Gated-SFT & LoRA & $-1.6\ [-2.4,{-}0.8]$ & 28.0\% & \multicolumn{2}{c}{not tested}\\
\bottomrule
\end{tabular}
\end{adjustbox}
\end{table}

The 1.7B results show the same pattern as the 8B.
\begin{itemize}
\item \emph{Steered-SFT} saves 26--33\% of tokens but loses 2.3--3.6 pp, so it
does not pass the $-2$ pp rule.
\item \emph{DPO-steered} keeps accuracy: it passes the rule on both sets with
both tuning types, and saves 15--19\% of tokens. On AIME/AMC, the full-tuned
version changes accuracy by $+2.9$ pp [$-0.4,{+}6.2$] and saves 9.0\%.
\item \emph{DPO-natural} saves more (23--25\%). Its full-tuned version also
passes the rule; its LoRA version does not pass it on either set. Compared
directly with DPO-natural, DPO-steered is 6--11\% longer. So, as on the 8B,
steered data does not improve DPO.
\end{itemize}
LoRA and full fine-tuning give the same verdict on every comparison of steered
with natural data.

SFT$\to$DPO does not give the best student. Its DPO stage recovers about half
of the SFT loss on levels 4 and 5 (+2.5 and +3.4 pp), but it also makes levels
2 and 3 longer (+11\% and +14\%). Against base, it still loses 1.8 pp
[$-2.6,{-}1.1$] on fresh1000. Gated-SFT has the smallest loss among the SFT
students that shorten, but it also does not pass the rule. We did not run
SFT$\to$DPO on the 8B: the 8B solves about 96\% of its training questions, so
too few hard questions have both a right and a wrong trace to build DPO pairs.
Offline training therefore supports reuse of the intervention, but it does not
solve the hard-problem limit.

\section{Implementation and reproduction}
\label{sec:evidence}
\paragraph{Steering hook.} We steer inside vLLM \cite{kwon2023vllm}. vLLM keeps
the hidden state and the residual as separate tensors. So the hook edits their
sum, which is the full residual stream. On mixture-of-experts layers, the edit
happens after the expert outputs are combined. The hook stores only the vector
and the schedule. It does not call the AV or the AR, and it does not change any
weights. Site and replay audits confirm that injections happen at eligible
tokens.

\paragraph{NLA checkpoints.} Every experiment uses one NLA checkpoint pair,
saved after one epoch of joint training: the AV \path{checkpoints/av_rl_ep1}
and the AR \path{checkpoints/ar_rl_ep1}. Both read layer 20, about 70\% of the
depth of Qwen3-1.7B, and the AV input has norm 1,000. Settings are in
\path{src/reasoning_attention/config.py}, in the launchers
\path{scripts/train_sft.sh} and \path{scripts/train_grpo.sh}, and in the
\path{nla_meta.yaml} file of each checkpoint. Code is at
\url{https://github.com/Strongich/doomloop-interp}; the repository also links
to the checkpoints.

\paragraph{NLA training.} Training has two stages.
\begin{enumerate}
\item \emph{Warm start.} We use about 500,000 positions from 100,000
Ultra-FineWeb documents. We split the documents into two disjoint halves: one
half trains the AV (249,798 rows), and the other trains the AR. A local
Qwen3.8-27B writes the summary of each position. Both models are fully
fine-tuned for one epoch, with a cosine learning rate that falls from
$2\times10^{-5}$ to $2\times10^{-6}$. The AV warm start uses 5\% warmup and an
effective batch of 256 (957 steps).
\item \emph{Joint training.} We use 393,439 new positions (about 400k) from web
text (Ultra-FineWeb) and chat (WildChat). The AV is trained with GRPO: it
samples 8 explanations per state at temperature 1, its reward is
$-\log\lVert h-\mathrm{AR}(e)\rVert^2$, and a KL penalty of 0.01 keeps it close
to the warm-start AV. At the same time, the AR is trained on the squared error
between normalized vectors. Each step uses 64 states $\times$ 8 samples and
learning rate $9.97\times10^{-6}$. One epoch is 6,147 steps on two A100 GPUs.
This is smaller than the released reference run on Qwen2.5-7B
\cite{frasertaliente2026nla}. That run used 500k web positions, batches of
128 states $\times$ 8 samples and 4,199 steps (about 4.3M samples, against our
3.1M), and reached FVE 0.752.
\end{enumerate}

\paragraph{Reconstruction metric.} The FVE of Eq.~\ref{eq:fve} is the
\texttt{fve\_nrm} metric of the reference code \cite{frasertaliente2026nla}.
It scales each state and each reconstruction to unit length, and it does not
scale the mean back to unit length. We sample three explanations per state at
temperature 1 and average the FVE over the three samples.

\paragraph{The deployed direction.} To build $N$, the AV writes explanations
with greedy decoding (no sampling) and at most 300 new tokens (the script
default). The deployed $N$ is \path{data/pool/dir_A_1trace.pt}. The policy and
comparator manifests lock its SHA256,
\texttt{6efc847671486f95add72341f2d9a2c0\allowbreak
b4d34497deb73a487b1bab3a31a1ad44}.

\paragraph{Two different scales.} The AV input has norm 1,000 because the NLA
was trained with that input scale (Section~\ref{method:interface}). The
steering strength $\alpha$ is unrelated to it: $\alpha$ scales the added vector
relative to the current residual of the steered model (Eq.~\ref{eq:steer}).

\paragraph{Paragraph boundaries and delay.} A boundary token is any vocabulary
token whose text contains at least two newline characters, for example
\texttt{\textbackslash n\textbackslash n}, \texttt{.\textbackslash n\textbackslash n},
or \texttt{)\textbackslash n\textbackslash n}. It must be generated after the
prompt and before the first \texttt{</think>}. The generated \texttt{<think>}
opener has index 1. It is never a site, because a single newline follows it.
So $d=0$ and $d=1$ give the same policy.

\paragraph{Doubt rules in detail.} The 13 doubt markers are:
``Wait'', ``Hmm'', ``Hold on'', ``Actually'', ``Alternatively'', ``But'',
``Maybe'', ``However'', ``Let me reconsider'', ``Let me double-check'', ``Let
me verify'', ``That doesn't seem right'', and ``That can't be right''. Two
rules use them.
\begin{enumerate}
\item \emph{Counting doubt paragraphs.} We take the thinking text, split it at
blank lines, and skip the first block. For each later block, the first sentence
ends at the first ``.'', ``!'', or ``?'' that is followed by whitespace. The
block counts if any marker occurs in that sentence as a whole word, ignoring
case.
\item \emph{Choosing donor sites.} We split the thinking text at blank lines.
We keep a break if the first 60 characters after it (leading whitespace
removed) start with a marker, ignoring case. The full prefix of a donor site
(prompt and response) must be at most 4,000 tokens long.
\end{enumerate}

\paragraph{The captured token.} At a donor site, the captured token is the last
token that ends at or before the break, so it is the token that carries the two
newlines. Steering uses the same token. In Qwen3 the last punctuation of a
paragraph usually joins the break in one token, such as
\texttt{.\textbackslash n\textbackslash n}. So ``the last token of the
paragraph'' and ``the break token'' were the same token at 295 of 297 doubt
sites in an earlier check. When the break is a separate
\texttt{\textbackslash n\textbackslash n} token, as at $t=1809$ below, we use
that newline token.

\paragraph{Donor sites.} The 35 donor sites are at response tokens 603, 635,
648, 672, 786, 1311, 1340, 1460, 1809, 1851, 1876, 1887, 2039, 2074, 2082,
2171, 2199, 2493, 2530, 2610, 2640, 2661, 2766, 3020, 3116, 3147, 3178, 3190,
3341, 3407, 3424, 3670, 3697, 3724, and 3769. Indices count from zero within the
saved response. In Figure~\ref{construction:trace}, the token at $t=1809$ is
\texttt{\textbackslash n\textbackslash n} (ID 271), and the token at $t=1851$ is
\texttt{.\textbackslash n\textbackslash n} (ID 382). Rebuilding $N$ from the
saved difference vectors of these sites with Eq.~\ref{eq:edit} gives a
bit-identical vector.

\paragraph{The example explanation.} The explanation in
Figure~\ref{fig:explanation} was saved when we rebuilt the direction ($N'$)
with the same AV and decoding. The saved data of the deployed $N$ keep only the
difference vectors, not the explanations.

\end{document}